\documentclass[times,twocolumn,final,authoryear]{elsarticle}

\usepackage{ycviu}
\usepackage{framed,multirow}

\usepackage[edges]{forest}
\usepackage{nicematrix}
\usepackage{siunitx}
\usepackage{booktabs}

\usepackage{amssymb}
\usepackage{latexsym}

\usepackage{url}
\usepackage{xcolor,colortbl}

\definecolor{gray}{rgb}{.75,.75,.75}

\definecolor{newcolor}{rgb}{.8,.349,.1}

\usepackage{amssymb}
\usepackage{makecell}
\usepackage{gensymb}
\usepackage{textcomp}
\usepackage{longtable}
\usepackage{hyperref}
\usepackage{tikz}
\usetikzlibrary{trees}
\usepackage{tikz}
\usepackage{soul}

\usepackage[absolute,overlay]{textpos}

\usepackage{caption}
\usetikzlibrary{arrows.meta, shapes.geometric, calc, shadows}
\colorlet{linecol}{black!75}

\usepackage{forest}
\usepackage{newfloat}
    \DeclareFloatingEnvironment[fileext=lod]{Figure} 

\journal{Computer Vision and Image Understanding}
\begin{document}

\begin{frontmatter}

\title{A survey on RGB-D datasets}

\author[1]{Alexandre \snm{Lopes}\corref{cor1}} 
\cortext[cor1]{Corresponding author.} 
\ead{alexandre.lopes@ic.uncamp.br}

\author[2,3]{Roberto \snm{Souza}}
\ead{roberto.medeirosdeso@ucalgary.ca}

\author[1]{Helio \snm{Pedrini}}
\ead{pedrini@unicamp.br}

\address[1]{Institute of Computing, University of Campinas, Brazil}
\address[2]{Department of Electrical and Computer Engineering, University of Calgary, Canada}
\address[3]{Hotchkiss Brain Institute, University of Calgary, Canada}

\received{1 May 2013}
\finalform{10 May 2013}
\accepted{13 May 2013}
\availableonline{15 May 2013}
\communicated{S. Sarkar}

\begin{abstract}
RGB-D data is essential for solving many problems in computer vision. Hundreds of public RGB-D datasets containing various scenes, such as indoor, outdoor, aerial, driving, and medical, have been proposed. These datasets are useful for different applications and are fundamental for addressing classic computer vision tasks, such as monocular depth estimation. This paper reviewed and categorized image datasets that include depth information. We gathered 231 datasets that contain accessible data and grouped them into three categories: scene/objects, body, and medical. We also provided an overview of the different types of sensors, depth applications, and we examined trends and future directions of the usage and creation of datasets containing depth data, and how they can be applied to investigate the development of generalizable machine learning models in the monocular depth estimation field.

\end{abstract}


\end{frontmatter}

\begin{textblock*}{18cm}(1.4cm,0.5cm) 
\Large{Final Open Access Version at: \url{https://doi.org/10.1016/j.cviu.2022.103489}}

\end{textblock*}


\section{Introduction}
\label{sec:sample1}

Depth is a critical information for many computer vision and image analysis applications. For example, it has been applied for tasks such as synthetic object insertion in computer graphics~\citep{luo2020consistent}, robotic grasping~\citep{lenz2015deep} automatic 2D to 3D conversion in film~\citep{xie2016deep3d}, robot-assisted surgery~\citep{stoyanov2010real}, and autonomous driving~\citep{5940562}. 

Despite using depth sensors that capture the distance information, researchers also use stereo vision matching to infer it, especially for its condensed size and cost. Lately, deep learning methods are being used to produce more precise and dense depth maps. For example, they can improve finer-grained details~\citep{miangoleh2021boosting}, produce dense maps from sparse inputs~\citep{8374553}, and refine depth for mirror surfaces~\citep{tan2021mirror3d}.

An important field of study for depth is monocular depth estimation, especially because it does not require using depth sensors, reducing the size and cost of computer vision systems' setups. Also, it can be applied to existing monocular systems, that comprise the majority of image capturing systems available. For instance, Light Detection And Ranging (LiDAR) scanners usually cost thousands of dollars, and their cost and weight can be impractical for many small drone applications.

As a result of the extensive range of applications of depth, a considerable number of datasets include distance measurements of points of the scene they acquire. These datasets are collected using different sensors in distinct scenes for applications such as Simultaneous Localization and Mapping (SLAM)~\citep{sturm12iros}, Reconstruction~\citep{dai2017scannet}, Object Segmentation~\citep{McCormac:etal:ICCV2017}, and Human Activity Recognition~\citep{zhang2016large}. With the increasing number and diversity of datasets, researchers were able to explore more generalistic forms of depth estimation, leading to techniques focused on zero-shot cross-dataset depth estimation~\citep{li2018megadepth, xian2020structure, Ranftl2021, Ranftl2020}. The idea is to produce powerful methods able to estimate depth for in-the-wild scenes, increasing the range of applications for depth estimation.

The main contribution of this paper is to categorize and summarize the existing datasets with depth data. We propose a survey that can be used by researchers of both individual applications and general systems. While there are good reviews of RGB-D datasets~\citep{firman2016rgbd, cai2017rgb}, the most recent one was published in 2017, and datasets have evolved both in complexity and size since then. Our survey presents a comprehensive literature review on more than 200 publicly available datasets included from an initial list of more than 300 datasets. Nearly half of the public datasets were published in 2017 or after, therefore, not included in any other review. We also made this work available on a website\footnote{\url{www.alexandre-lopes.com/rgbd-datasets}} to facilitate the filtering by application, scene type, sensor, and year.

The remainder of this paper is structured as follows. In the next section, we discuss and categorize depth sensors, explaining the main differences and applications for each category. In Section \hyperref[sec:datasets]{3}, we present the methodology used to perform the literature review. In Section \hyperref[sec:datasets]{4}, we present the datasets divided into categories, describing the most influential datasets for each category and presenting the rest in tables. In Section \hyperref[sec:discussion]{5}, we present tendencies and discuss future directions for RGB-D data usage. Finally, we provide a summary of the field and discuss how the area is evolving in Section~\hyperref[sec:conclusion]{6}.

\section{Sensors}
\label{sec:sensors}

Range (or depth) data is crucial for understanding the 3D scene projected onto a 2D plane forming an image. There are multiple ways to obtain such information, either using a depth sensor or estimating depth. A depth sensor is a device that provides the distance from the sensor to an element in the scene, although it is possible to collect distance information using two or more RGB cameras from a scene. We define as Stereo Camera Sensing, all systems formed by two or more cameras. Therefore, light field cameras are also included here.

\begin{figure*}[!t]
\centering
\begin{tikzpicture}[picture format/.style={inner sep=1pt, scale=1.35}]

  \node[picture format]                   (A1)               {\includegraphics[width=1in]{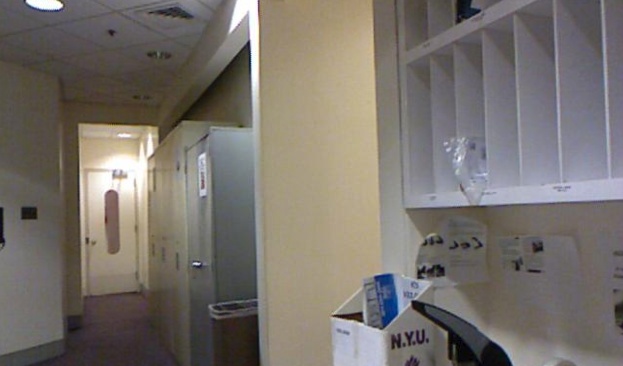}};
  
  \node[label=below:{\small (a)}, picture format,anchor=north]      (B1) at (A1.south) {\includegraphics[width=1in]{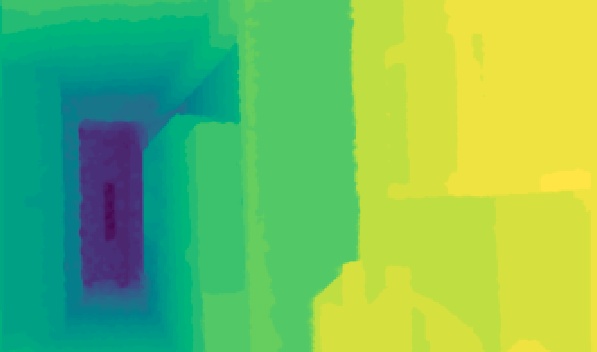}};

  \node[picture format,anchor=north west] (A2) at (A1.north east) {\includegraphics[width=1in]{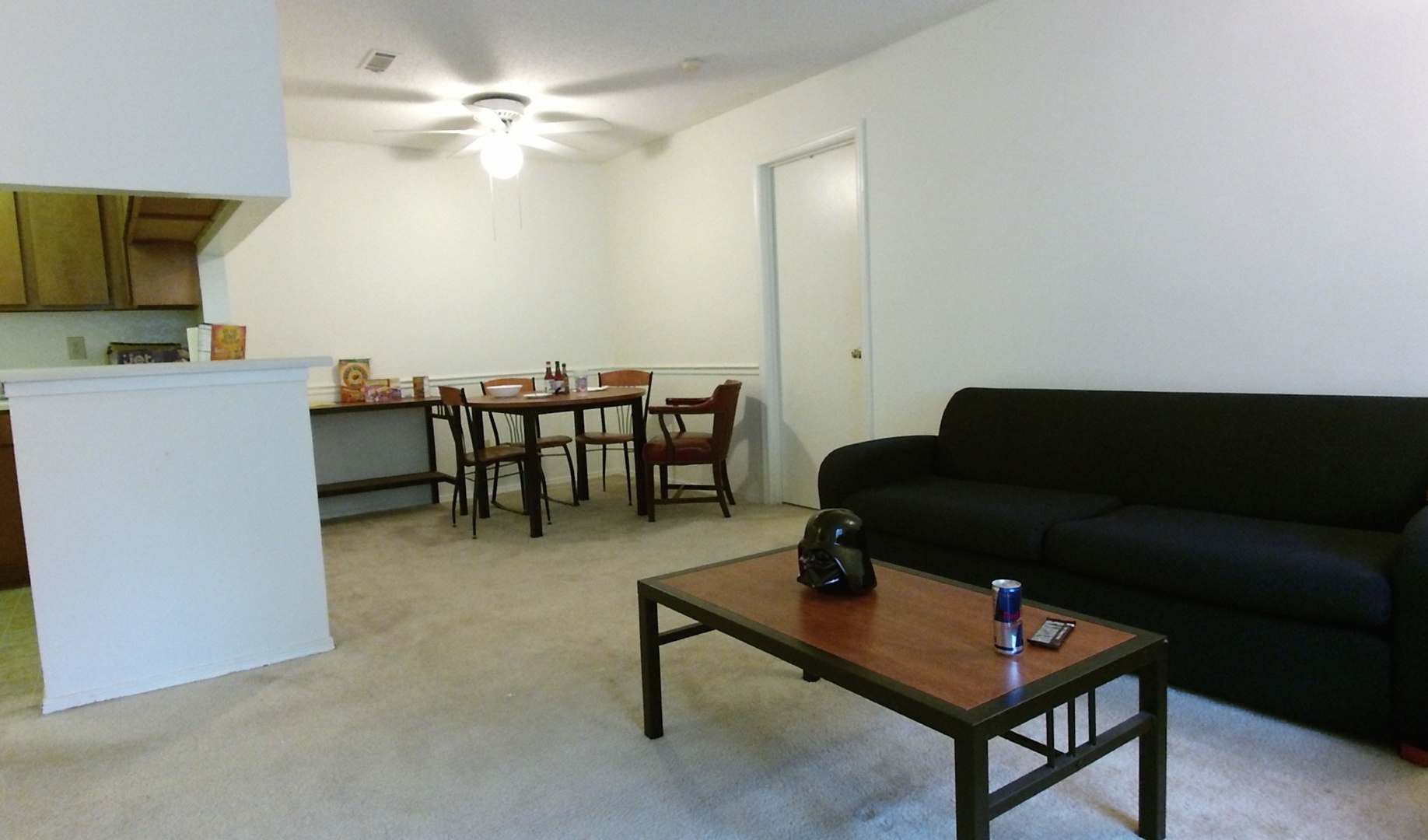}};
  \node[label=below:{\small (b)},picture format,anchor=north]      (B2) at (A2.south)      {\includegraphics[width=1in]{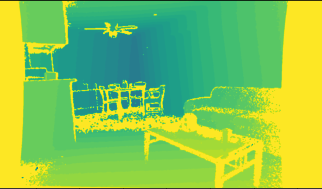}};

  \node[picture format,anchor=north west] (A3) at (A2.north east) {\includegraphics[width=1in]{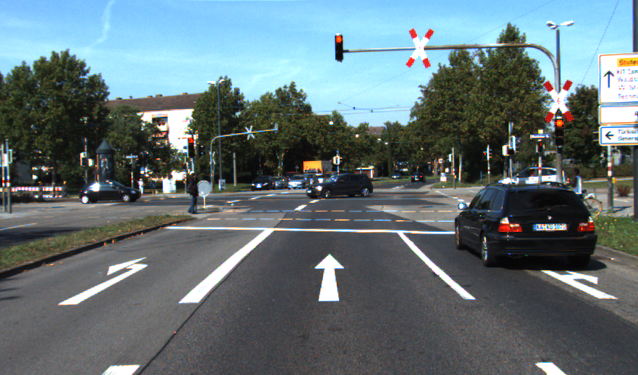}};
  \node[label=below:{\small (c)},picture format,anchor=north]      (B3) at (A3.south)      {\includegraphics[width=1in]{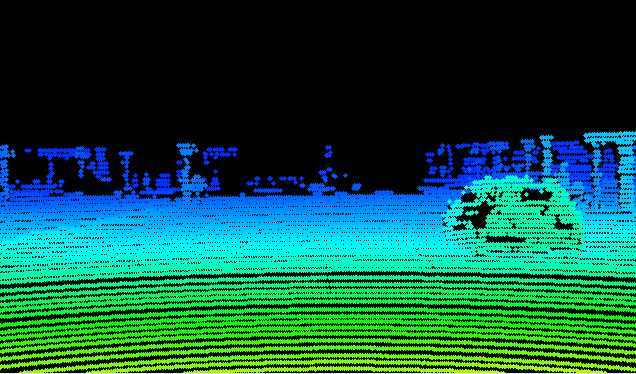}};

  \node[picture format,anchor=north west] (A4) at (A3.north east) {\includegraphics[width=1in]{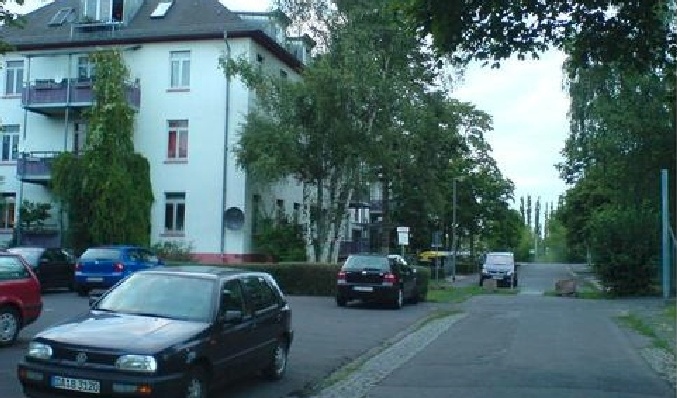}};
  \node[label=below:{\small (d)},picture format,anchor=north]      (B4) at (A4.south)      {\includegraphics[width=1in]{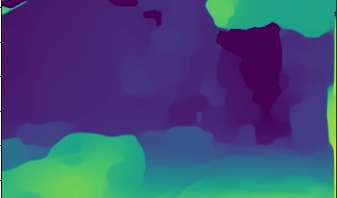}};

\end{tikzpicture}
\caption{Examples of depth data with image (first row) and depth (second row) of the following sensors: \textbf{(a)} Structured Light from NYUv2~\citep{10.1007/978-3-642-33715-4_54}, \textbf{(b)} TOF from AVD~\citep{active-vision-dataset2017}, \textbf{(c)} LiDAR from KITTI~\citep{Geiger2013IJRR}, and \textbf{(d)} Stereo Camera Sensing from ReDWeb~\citep{Xian_2018_CVPR}, where the authors compute correspondence maps by using optical flow.}
\label{fig:depthexamples}

\end{figure*}

Previously, authors proposed distinct divisions for the types of sensors~\citep{Fisher2008, Choi2019}. In this survey, we use a categorization of depth sensors inspired by~\citet{Choi2019}'s work. We divide the sensors into the following categories: Structured Light, Time-of-Flight (TOF),  Light Detection and Ranging (LiDAR), and Stereo Camera Sensing. We display examples of each category in Figure~\ref{fig:depthexamples}.

Ultrasonic and Radar sensors also produce distance information, but they are out of the scope of this work because they are rarely used to produce depth information associated with RGB data. We detail each one of the sensors categories in the following sub-sections and show the differences of these types and possible application scenes in Table~\ref{Tabela6}.

\begin{table*}[!ht]
\caption{Sensors overview comparing the usual application, distances and sparsity for each type of sensor.}
\label{Tabela6}
\resizebox{\textwidth}{!}{%
\setlength{\tabcolsep}{2.6em} 
\begin{tabular}{cccc}
\toprule[1.5pt]

\textbf{Type} &  \textbf{Typical Application Scenes} & \textbf{Typical Distance Usage} & \textbf{Sparsity}\\
\midrule[1.25pt]

\rowcolor{gray!25}
Structured Light & Indoor &  Close Distances (0-10m) & Dense Map\\
TOF & Indoor &  Close Distances (0-10m) & Dense Map\\
\rowcolor{gray!25}
LiDAR & Aerial, Street, Outdoor, Indoor & Medium/Large Distances (10-1000m)& Sparse Map\\
Stereo Camera Sensing & Street, Outdoor, Indoor & Close/Medium Distances (0-100m) & Dense Map \\
\bottomrule[1.5pt]
\end{tabular}
}
\end{table*}

\subsection{Structured Light}

Structured Light sensors (also called Active Stereo sensors) rely on a projector of light captured by a camera. The simplest way to achieve such a goal is to project a point with a device and capture this point in the scene with the camera. The depth of this point can be measured by a technique called Triangulation. For estimating depth, it is necessary to find the position of the projected point in the image plane, have the distance between the camera and the light projector, the camera's internal parameters, and the position in space of the projector. With this information, it is possible to create a triangle and calculate the height of the triangle formed by the camera, projector, and illuminated scene point to determine the distance. The strategy of projecting points would be slow in practice since it is necessary to project a point for every position that is represented as a pixel in the image. 

A more efficient strategy is to project the light as a stripe that associated with different coding strategies, such as the Binary Coded Structured Light strategy, can reduce the number of frames necessary to produce a full depth map. It can also be coded with RGB lights. Details about different codification strategies are discussed by~\citet{SALVI2004827}. 

Most Structured Light sensors do not work under direct sunlight since they rely on light projection in a scene. Therefore, they are usually suitable for indoor scene applications. Researchers have proposed strategies to overcome challenging light conditions~\citep{o2015homogeneous}, and now these sensors appear in smartphones for face identification systems for both indoor and outdoor scenes. They typically have a low range limit, not going further than 10 meters. Examples of this type of sensor include Matterport, Kinect~v1, and RealSense SR300 cameras.

\subsection{Time-of-Flight}

TOF sensors estimate the distance of an object in the scene to a sensor by measuring the time it takes for an emitted light to be received by the sensor. Therefore, TOF sensors rely on the time that a light wave takes to go to a point in a scene and to be reflected to a sensor. The concept is barely the same as the Ultrasonic and Radar Sensors, but here light is used as the emitted signal.

There are multiple strategies for capturing the time-of-flight of light. The most straightforward strategy is using a technique called Pulse Modulation, where a very fast pulse of light is emitted and then received by the sensor. The time delay between the emitted light pulse and the received light pulse is used to compute the distance of the object in the scene. Continuous-Wave Modulation is another strategy, where the light is modulated by its intensity, and the distance is measured by calculating the shift in phase of the original emitted light and the received light.

TOF sensors generally are compromised under strong sunlight conditions~\citep{kazmi2012plant}, making this sensor more commonly applied to indoor scenes. Existing studies try to overcome the effect under intense background light~\citep{buttgen2008robust} and to reduce the measurement uncertainty under such conditions. Examples of this type of sensor include Kinect~v2 (Xbox One sensor), SoftKinetic DS 325, and RIEGL VZ-400.

\subsection{LiDAR}

LiDAR sensors use the same idea of measuring the time that an emitted light is received by a sensor, but they rely on one or multiple laser beams (concentrated light) to produce depth measurements of points in the scene, and the device usually has a rotating mirror to generate 360\textdegree{} scans of a scene. Hence, LiDAR sensors produce point clouds of a scene, not a dense depth map of it. They rely on focused laser beams, which allow them to collect distance measurements as far as a few kilometers. LiDAR sensor models have different specifications (e.g., resolution, scans per second, and distance accuracy), and some scans are built in a multilayer (multiple laser beams) configuration, allowing them to measure not only in a 360\textdegree{} plane of the sensor but in 3D.

LiDAR measurement accuracy is usually independent of distance, although some models can fail in adverse weather conditions, such as dense fogs and turbulent snow~\citep{jokela2019testing}. Each LiDAR point also includes the intensity measurements, which can be interpreted as a measurement of reflectivity of the point that the light hit. This value is suitable for many applications, such as vegetation cover understanding and tunnel damage detection~\citep{kashani2015review}, giving LiDAR additional information that other types of sensors do not produce.

LiDAR sensors emit light; therefore, they work in difficult lighting conditions, such as dark environments. They are suitable for indoor and outdoor application scenes, but the available models are usually limited to specific applications, such as aerial measurements, outdoor/driving applications, and small indoor spaces depth estimation. Examples of such types of sensors include Velodyne Sensors, Faro Focus 3D Laser, and SICK LMS-511.

\subsection{Stereo Camera Sensing}

We define here Stereo Camera Sensing (SCS) as any system formed by two or more image sensors or lenses used to produce a Depth Map of a scene. Hence, simplistic pairs of cameras and complex light field systems composed by multiple microlenses are both identified in the same category. A straightforward strategy to measure depth from two or more cameras is Triangulation. The Triangulation idea is the same as applied in Structured Light sensors, but using a camera instead of a projector. The idea is that finding the position of a pixel in the image plane of camera $A$ projected from a point $P$ in the space, and the position of a pixel projected by the same point $P$ in camera $B$, it is possible to find the depth of that point in a scene with the intrinsic parameters of the camera. After finding both lines projected in both cameras from point $P$, it is only necessary to know the distance between the two cameras (baseline distance) and internal parameters of the cameras to know the depth of the point $P$.

A limitation of this strategy occurs when the point of interest has no texture. For instance, it is practically impossible to determine which point of a smooth painted wall observed in the image projected by camera $A$ is equivalent to the image projected by camera $B$. Therefore, it is difficult to determine a point's depth with acceptable accuracy without the correspondence of the pixels in both image planes. Recently, Deep Learning based methods have tried to address this limitation, increasing the accuracy of the estimation~\citep{zbontar2015computing}. Examples of such types of sensors include light field cameras and ZED cameras.

\section{Methodology}
\label{sec:methodology}
A literature review should synthesize previous knowledge, identify biases and gaps in the literature~\citep{Rowe2014}. Since our study aims to describe, categorize, and identify future trends for RGB-D datasets, we defined a non-conventional methodology to find the related papers. Instead of defining search terms to find the papers directly, we collected datasets using backward snowballing. The premise is that many datasets containing depth data do not have depth estimation as their primary goal, as in KITTI Dataset~\citep{Geiger2013IJRR}. Therefore, defining search strings that could find depth datasets using generalist terms would result in numerous false-positive results. For instance, the search string \texttt{RGB-D OR Depth AND Dataset} searching in abstract, keywords, or title brings more than 23 thousand results in Scopus. Moreover, if we define a complex composed search string to filter the results, we would miss many datasets in the search.

As monocular depth estimation, salient object detection, and action recognition are prominent fields in the area, we defined the following search string to perform backward snowballing: \texttt{(("single image" OR monocular) AND depth AND estimation) OR (("Salient Object Detection" OR "Action Recognition") AND RGB-D)}. The terms ``monocular'' and ``single image'' are applied mainly for monocular depth estimation but are also used for stereo trained systems, depth completion, and other applications. We conducted the review in Scopus and Google Scholar search engines. In Scopus, we revised all papers from January 1st, 2016, through August 31st, 2021. From Google Scholar, we followed the same dates, but we also included a stop criterion. If we found one search page without relevant items, we would end the year's search. The inclusion of Google Scholar is justified because many relevant papers are published in arXiv. Consequently, those could also be included in this work.

The exclusion and inclusion criteria for papers are defined in Table~\ref{TabelaIncexc}. These criteria are applied to the papers found using the previous search term. After excluding papers, backward snowballing was applied to find the datasets used/described by the remaining works. Initially, we reviewed 2,119 papers, which led to 374 dataset candidates. We also applied an exclusion criterion to these candidates, and only papers with active project websites, contact information to download the dataset, or direct download link were included. Hence, the final list of datasets to be included was reduced to 231 datasets.

\begin{table}[!t]
\caption{Inclusion and Exclusion Criteria.}
\label{TabelaIncexc}
\resizebox{\columnwidth}{!}{
\setlength{\tabcolsep}{1.0em} 
\begin{tabular}{cc}
\toprule[1.5pt]

\textbf{Criterion} &  \textbf{Category} \\
\midrule[1.25pt]

\rowcolor{gray!25}
Papers that discuss depth estimation & Inclusion \\
\makecell[c]{Papers using depth sensors, stereo image \\ sensing, or synthetic data} & Inclusion      \\
\rowcolor{gray!25}
Papers not written in English & Exclusion \\
Papers exclusively using private datasets & Exclusion \\
\rowcolor{gray!25}
Papers not presenting minimal evidence of valid results & Exclusion\\
Duplicated paper/report. We kept  the most
complete one & Exclusion\\
\bottomrule[1.5pt]
\end{tabular}
}

\end{table}

\section{Datasets}
\label{sec:datasets}

\begin{figure*}[!t]
\centering
        \pgfkeys{/forest,
          my rounded corners/.append style={rounded corners=2pt},
        }
        \begin{forest}
          for tree={
              line width=1pt,
              draw=linecol,
              fit=rectangle,
              edge={color=linecol, >={Triangle[]}, ->},
              where level=0{%
                l sep+=5pt,
                calign=child,
                calign child=2,
                align=center,
                my rounded corners,
                for descendants={%
                  calign=first,
                },
              }{%
                where level=1{%
                  my rounded corners,
                  align=center,
                  parent anchor=south west,
                  tier=three ways,
                  for descendants={%
                    child anchor=west,
                    parent anchor=west,
                    align=left,
                    anchor=west,
                    edge path={
                      \noexpand\path[\forestoption{edge}]
                      (!to tier=three ways.parent anchor) |-
                      (.child anchor)\forestoption{edge label};
                    },
                  },
                }{}%
              },
          }
[Dataset,
    [Scene/Objects
        [{SLAM, Odometry, \\ or Reconstruction}
        [{Segmentation or Other \\ Extra Information}
        [{Depth Data Only}
        [{Other}]
        ]]]]
    [Body 
        [Human Activities 
        [Gestures (Partial body)]]]
    [Medical
        ]
]
        \end{forest}
    \caption{Taxonomy for RGB-D datasets.}
    \label{fig:tree}
\end{figure*}
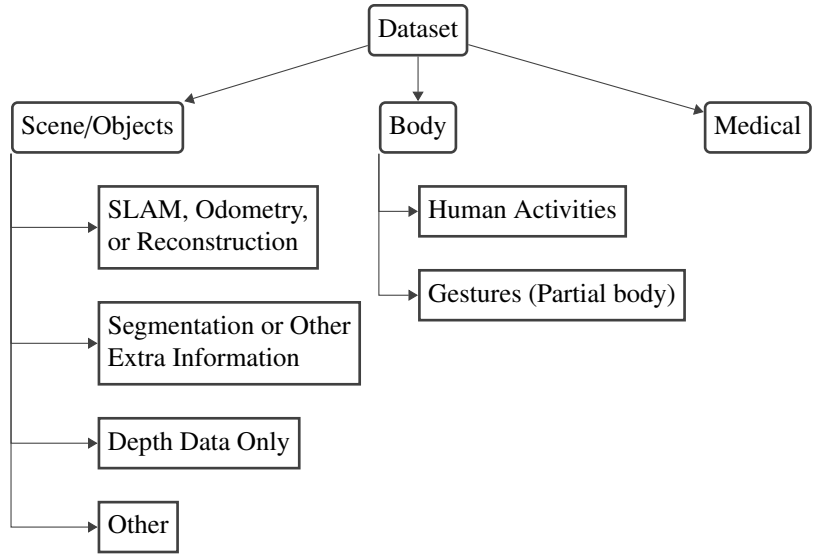

In recent years, many datasets have been created using the sensors or stereo vision sensing presented in the previous section. In addition to datasets using real data, this paper also includes datasets containing synthetic data. These were created mainly by simulation systems and often presented extra data such as semantic segmentation and 3D object detection bounding boxes. We divided the selected datasets into three different categories and six different sub-categories representing different application areas. The taxonomy tree is available in Figure~\ref{fig:tree}.

The categories represent the intended application of the dataset. In the first level, we identify datasets that are mainly interested in Scenes/Objects, Human Body, or Medical Applications. The following sub-sections explore each application area, and list all of them in each sub-category's table. We also detail three, two or one datasets for each sub-category, based on the total number of datasets of each sub-category. If we detail three papers, the two first ones are the most cited papers that contain complementary scenarios. For example, KITTI Dataset and ScanNet Dataset contain street and indoor scenes, respectively. The third paper is the most cited paper published in 2017 or later. If we detail two papers, these are the most cited ones that contain complementary scenarios, and if we detailed one paper, it is the most cited in the sub-category.




\subsection{Scene/Objects}

In this category, we grouped all datasets generally intended to expose scenes, individual objects, or groups of objects containing or not humans.. Therefore, datasets that reconstruct scenes/objects, segment elements of a scene, salient objects using depth, and contain exclusively depth maps are sub-categorized here. We created an ``Other'' sub-category to accommodate datasets that did not fit into these previous sub-categories.

Some papers explore multiple applications, primarily synthetic datasets, since they can create reconstruction and segmentation data directly using simulation environments. These papers are presented in one of their application areas to reduce redundancy. The only exception is for datasets of ``SLAM, Odometry, or Reconstruction'' and ``Segmentation or Other Extra Information'' sub-categories that are presented together in Table~\ref{TabelaSOR_SOE}, since this combination is very frequent for datasets.

\subsubsection{SLAM, Odometry, or Reconstruction}

This sub-category contains multiple types of applications, however, all of them have a common characteristic: they present extra information that makes possible to recreate in any detail level, a 3D scene. For SLAM and odometry related papers, they typically present camera pose information, giving position and orientation of the capturing apparatus of each frame/image. We treated odometry differently from SLAM since odometry essentially aims to estimate the path of the camera, and SLAM tries to obtain a consistent trajectory and scene map of the camera~\citep{yousif2015overview}.

All collected datasets that contain data exclusively for SLAM, Odometry, or Reconstruction are shown in Table~\ref{TabelaSOR}. In general, applications of indoor scenes focus on reconstruction, and external scenes (such as driving scenes) focus on SLAM/odometry. Table~\ref{TabelaSOR_SOE} also contains datasets of this sub-category, however, with extra annotated information such as semantic segmentation data. 
Some of the most cited datasets in the field include:
\newline

\begin{table*}[!t]

\caption{Datasets of ``SLAM, Odometry, or Reconstruction'' sub-category}
\label{TabelaSOR}
\resizebox{\textwidth}{!}{%

\begin{NiceTabular}{lllllllll}
\toprule[2pt]
                                                                           \textbf{Dataset Name} & \textbf{Ref.} & \textbf{Year} &     
\textbf{Scene Type} &  \textbf{Sensor Type} &                                              \textbf{Sensor Name} &                                                           \textbf{Data Modalities} &                                         \textbf{Extra Data} &                                                                                                          \textbf{Images/Scenes} \\
\midrule[1.5pt]

 \rowcolor{gray!25}
                                          GL3D &                    \cite{shen2018mirror} &           2018 &                                                             Aerial &                                               SCS &                                                                                     Stereo Camera &                                                              Color, Depth &                                               - &                                 \makecell[l]{543 Scenes (125623 \\ Images)} \\

                                   ApolloScape &               \cite{wang2019apolloscape} &           2020 &                                                            Driving &                                             LiDAR &                                                    \makecell[l]{Velodyne HDL-64E \\ S3} &                                 \makecell[l]{Color, Depth, GPS, \\ Radar} &                                               - &                                    \makecell[l]{155 Min With 93k \\ Frames} \\
 \rowcolor{gray!25}                                          KAIST &                 \cite{ jjeong-2019-ijrr} &           2019 &                                                            Driving &                                        SCS, LiDAR &                                     \makecell[l]{Velodyne VLP-16, SICK \\  LMS-511, Stereo Camera} &                        \makecell[l]{Color, Depth, GPS, \\ IMU, Altimeter} &                                               - &                                      \makecell[l]{19 Sequences (191 \\ Km)} \\
                                      RobotCar &               \cite{RobotCarDatasetIJRR} &           2016 &                                                            Driving &                                             LiDAR &                 \makecell[l]{2 X SICK LMS-151 2D \\ LiDAR, 1 X SICK \\ LD-MRS 3D LiDAR} & \makecell[l]{Color, Deph, GPS, \\ INS (Inertial \\ Navigation \\ System)} &                                               - &   \makecell[l]{133 Scenes (almost \\ 20M Images (from \\ Multiple Sensors)} \\
 \rowcolor{gray!25}                                   Malaga Urban &                  \cite{blanco2014malaga} &           2014 &                                                            Driving &                                        SCS, LiDAR &                                              \makecell[l]{2 SICK LMS, 3 HOKUYO,\\ Stereo Camera} &                                   \makecell[l]{Color, Depth, IMU, \\ GPS} &                                               - &                                                                15 Sequences \\
                               Omniderectional &                \cite{Schoenbein2014ICRA} &           2014 &                                                            Driving &                                        SCS, LiDAR &                                                  \makecell[l]{Velodyne HDL-64E, \\ Stereo Camera} &                                                              Color, Depth &                                               - &                                  \makecell[l]{152 Scenes (12607 \\ Frames)} \\
  \rowcolor{gray!25} \makecell[l]{Ford Campus Vision \\ And LiDAR} &                    \cite{pandey2011ford} &           2011 &                                                            Driving &                                        SCS, LiDAR &                                                  \makecell[l]{Velodyne HDL-64E, \\ Stereo Camera} &                                   \makecell[l]{Color, Depth, IMU, \\ GPS} &                                               - &                                                                 2 Sequences \\
                                     Karlsruhe &                      \cite{Geiger2011IV} &           2011 &                                                            Driving &                                               SCS &                                                                                     Stereo Camera &                                                            Color, GPS/IMU &                                               - &                                \makecell[l]{20 Sequences \\ (16657 Frames)} \\
 \rowcolor{gray!25}     \makecell[l]{Multi-FoV (Urban \\ Canyon )} &                  \cite{zhang2016benefit} &           2016 &                                                    Driving, Indoor &                                                 - &                                                                               Synthetic &                                                              Color, Depth &                                               - &                                                                 2 Sequences \\
                                            -- &                \cite{zeisl2013automatic} &           2013 &                                                   Driving, Outdoor &                                                N/A &                                                                        RGB-D Scans (N/A) &                                                              Color, Depth &                                               - &         \makecell[l]{13 Scenes (5 \\ Castle, 5 Church, 3 \\ Street Scenes)} \\
 \rowcolor{gray!25}                                     BlendedStereo Camera &                 \cite{yao2020blendedmvs} &           2020 &                                                        In-the-wild &                                                 - &                                                                               Synthetic &                                                              Color, Depth &                                               - &                                  \makecell[l]{113 Scenes (17k \\ Images)} \\
                                    Youtube3D &                  \cite{chen2019learning} &           2019 &                                                        In-the-wild &                                                 - &                                  \makecell[l]{Two Points \\ Automatically \\ Annotated} &                                    \makecell[l]{Color, Relative \\ Depth} &                                               - &                                                               795066 Images \\
 \rowcolor{gray!25}                                             -- &                           \cite{8425011} &           2019 &                                                        In-the-wild &            \makecell[l]{Structured Light, \\ TOF} &                                            \makecell[l]{Kinect V1, V2 And \\ Synthetic} &                                                              Color, Depth &                                               - &                                    \makecell[l]{10 Scenes (2703 \\ Frames)} \\
    \makecell[l]{4D Light Field \\ Benchmark} &                \cite{honauer2016dataset} &           2016 &                                                        In-the-wild &                                               - &                                            \makecell[l]{Light-field \\ (Synthetic)} &                                                              Color, Depth &                                               - &                                                                   24 Scenes \\
 \rowcolor{gray!25}     \makecell[l]{Habitat \\ Matterport (HM3D)} &           \cite{ramakrishnan2021habitat} &           2021 &                                                             Indoor &                                  Structured Light &                                                                         Matterport Pro2 &                                                              Color, Depth &                                               - &                                                                 1000 Scenes \\
                                      MilliEgo &                    \cite{lu2020milliego} &           2020 &                                                             Indoor &          \makecell[l]{Structured Light, \\ LiDAR} &        \makecell[l]{Intel D435i Depth, \\ Velodyne HDL-32E / \\ Velodyne Ultra \\ Puck} &                                 \makecell[l]{Radar, IMU, LiDAR, \\ Depth} &                                               - &         \makecell[l]{17 Distinct Floors \\ From 6 Different \\ Multistorey} \\
 \rowcolor{gray!25}                                            ODS &                       \cite{lai2019pano} &           2019 &                                                             Indoor &                                               SCS &                                             \makecell[l]{MiniPolar 360 \\ Camera (Stereo Camera)} &                                                              Color, Depth &                                     Normal Maps &                              \makecell[l]{6 Indoor Areas \\ (50k Images)} \\
                                         360D &              \cite{zioulis2018omnidepth} &           2018 &                                                             Indoor &                                  Structured Light &                                        \makecell[l]{Synthetic And \\ Matterport Camera} &                                                              Color, Depth &                                               - &                \makecell[l]{12072 Scanned \\ Scenes And 10024 CG \\ Scenes} \\
 \rowcolor{gray!25}                                      PanoSUNCG & \cite{DBLP:journals/corr/abs-1811-05304} &           2018 &                                                             Indoor &                                                 - &                                                                               Synthetic &                                                              Color, Depth &                                               - &                                  \makecell[l]{103 Scenes (25k \\ Images)} \\
                                         CoRBS &             \cite{wasenmueller2016corbs} &           2016 &                                                             Indoor &                                               TOF &                                                                               Kinect V2 &                                                              Color, Depth &                                               - &                            \makecell[l]{4 Scenes (9 Hours Of \\ Recording)} \\
 \rowcolor{gray!25}                                      EuRoC MAV &                     \cite{Burri25012016} &           2016 &                                                             Indoor &                                          TOF, Stereo Camera &                                    \makecell[l]{Vicon Motion \\ Capture, Leica \\ MS50} &                                                         Color, Depth, IMU &                                               - &                                                                   11 Scenes \\
           \makecell[l]{Augmented \\ ICL-NUIM} &                    \cite{choi2015robust} &           2015 &                                                             Indoor &                                                 - &                                                                               Synthetic &                                                              Color, Depth &                                               - &                        \makecell[l]{4 Scenes (2 Living \\ Room, 2 Offices)} \\
 \rowcolor{gray!25}                                           Ikea &                    \cite{li2015database} &           2015 &                                                             Indoor &                                  Structured Light &                                               \makecell[l]{Kinect V1 And \\ PrimeSense} &                                                              Color, Depth &                                               - &                                                                    7 Scenes \\
                                        ViDRILO &               \cite{martinez2015vidrilo} &           2015 &                                                             Indoor &                                  Structured Light &                                                                               Kinect V1 &                                                              Color, Depth & \makecell[l]{Semantic Category \\ of the Scene} &                                 \makecell[l]{5 Sequences (22454 \\ Images)} \\
 \rowcolor{gray!25}                                       ICL-NUIM &               \cite{handa:etal:ICRA2014} &           2014 &                                                             Indoor &                                                 - &                                                                               Synthetic &                                                              Color, Depth &                                               - &                         \makecell[l]{8 Scenes (4 Living \\ Room, 4 Office)} \\
                                    MobileRGBD &          \cite{vaufreydaz2014mobilergbd} &           2014 &                                                             Indoor &                                               TOF &                                                                               Kinect V2 &                                                              Color, Depth &                                               - &                          \makecell[l]{3 Scenes (9.5 Hours \\ Of Recording)} \\
 \rowcolor{gray!25}                                RGBD Object  V2 &               \cite{lai2014unsupervised} &           2014 &                                                             Indoor &                                  Structured Light &                                                                               Kinect V1 &                                                              Color, Depth &                                               - &                                                                14 Sequences \\
                                            -- &                          \cite{MPMSP:14} &           2014 &                                                             Indoor &                                             LiDAR &                                                                     Faro Focus 3D Laser &                                                                     Depth &                                               - &                    \makecell[l]{40 Scenes (rooms \\ From Three \\ Offices)} \\
 \rowcolor{gray!25}                                RGB-D  7-Scenes &                   \cite{glocker2013real} &           2013 &                                                             Indoor &                                  Structured Light &                                                                               Kinect V1 &                                                              Color, Depth &                                               - &                           \makecell[l]{7 Scenes (500-1000 \\ Frames/scene)} \\
                                  Reading Room &                           \cite{6751168} &           2013 &                                                             Indoor &                                  Structured Light &                                                                     Asus Xtion Pro Live &                                                              Color, Depth &                                               - &                                                                     1 Scene \\
 \rowcolor{gray!25}                                       TUM-RGBD &                      \cite{ sturm12iros} &           2012 &                                                             Indoor &                                  Structured Light &                                                                               Kinect V1 &                              \makecell[l]{Color, Depth, \\ Accelerometer} &                                               - &                                                                39 Sequences \\
      \makecell[l]{IROS 2011 Paper \\ Kinect} &                           \cite{6094861} &           2011 &                                                             Indoor &                                  Structured Light &                                                                               Kinect V1 &                                                                     Depth &                                               - &                                                                27 Sequences \\
 \rowcolor{gray!25}                                             -- &                     \cite{zhou2013dense} &           2013 & \makecell[l]{Indoor, Isolated \\ Objects / Focussed \\ On Objects} &                                  Structured Light &                                                                     Asus Xtion Pro Live &                                                              Color, Depth &                                               - &                                                                    6 Scenes \\
                                            -- &                    \cite{meister2012can} &           2012 & \makecell[l]{Indoor, Isolated \\ Objects / Focussed \\ On Objects} &            \makecell[l]{Structured Light, \\ TOF} & \makecell[l]{KinectFusion \\ (Kinect V1) For Two \\ Scenes. Riegl \\ VZ-400 For Office} &                                                              Color, Depth &                                               - &                             \makecell[l]{2 Scenes: Statue \\ And Targetbox} \\

 \rowcolor{gray!25}                                           M\&M &                        \cite{ren2020suw} &           2020 &                                                    Indoor, Outdoor &                                               SCS &                                                                                     Stereo Camera &                                                              Color, Depth &                                               - &          \makecell[l]{4690 Sequences \\ (170k Frames) And \\ 130k Images} \\
          \makecell[l]{Mannequin \\ Challenge} &                             \cite{48104} &           2019 &                                                    Indoor, Outdoor &                                               SCS &                                                                                     Stereo Camera &                                                              Color &                                               - &                               \makecell[l]{4690 Sequences \\ (170k Frames)} \\                                            
 \rowcolor{gray!25}                                          Stereo CameraEC &               \cite{zhu2018multivehicle} &           2018 &                                                    Indoor, Outdoor &                                        SCS, LiDAR &                                                  \makecell[l]{Velodyne (LiDAR), \\ Stereo Camera} &                                                         Color, Depth, IMU &                                               - &                                                                 5 Sequences \\

                                          ETH3D &                   \cite{schoeps2017cvpr} &           2017 &                                                    Indoor, Outdoor &                                        SCS, LiDAR &                                 \makecell[l]{FaroFocus X 330 \\ (Laser Sensor), \\ Stereo Camera} &                                                              Color, Depth &                                               - &                                    \makecell[l]{25 High-res, 10 \\ Low-res} \\

 \bottomrule[2pt]

\end{NiceTabular}

}

\vspace{-0.3cm}
 \begin{flushright}\scriptsize Continue on Next Page \end{flushright}
\end{table*}

\begin{table*}[!t]

\resizebox{\textwidth}{!}{%

\begin{NiceTabular}{lllllllll}
\toprule[2pt]
                                                                           \textbf{Dataset Name} & \textbf{Ref.} & \textbf{Year} &     
\textbf{Scene Type} &  \textbf{Sensor Type} &                                              \textbf{Sensor Name} &                                                           \textbf{Data Modalities} &                                         \textbf{Extra Data} &                                                                                                          \textbf{Images/Scenes} \\
\midrule[1.5pt]

 \rowcolor{gray!25}                                   DiLigGent-MV &                       \cite{li2020multi} &           2020 &         \makecell[l]{Isolated Objects / \\ Focussed On \\ Objects} &                                               SCS &                                                                                     Stereo Camera &                                                                     Color &                                               - &                                                          5 Objects (scenes) \\
     \makecell[l]{A Large Dataset of \\Object Scans} &                          \cite{Choi2016} &           2016 &         \makecell[l]{Isolated Objects / \\ Focussed On \\ Objects} &                                  Structured Light &                                                     \makecell[l]{PrimeSense \\ Carmine} &                                                              Color, Depth &                                               - &                            \makecell[l]{Over 10k 3D Scans \\ Of Objects.} \\
 \rowcolor{gray!25}                                             -- &             \cite{zollhoefer2015shading} &           2015 &         \makecell[l]{Isolated Objects / \\ Focussed On \\ Objects} &            \makecell[l]{SCS, Structured \\ Light} &                                                                         \makecell[l]{PrimeSense, \\ Stereo Camera} &                                                              Color, Depth &                                               - & \makecell[l]{9 Scenes: 4 Scenes \\ Using PrimeSense, \\ 5 Scenes Using Stereo Camera} \\
                                       BigBIRD &                  \cite{singh2014bigbird} &           2014 &         \makecell[l]{Isolated Objects / \\ Focussed On \\ Objects} &                                  Structured Light &                                                    \makecell[l]{PrimeSense \\ Carmine 1.09} &                                                              Color, Depth &                                               - &                              \makecell[l]{600 Images (from \\ 125 Objects)} \\
 \rowcolor{gray!25}                                       Fountain &                     \cite{zhou2014color} &           2014 &         \makecell[l]{Isolated Objects / \\ Focussed On \\ Objects} &                                  Structured Light &                                                                     \makecell[l]{Asus Xtion\\ Pro Live} &                                                              Color, Depth &                                               - &                                                                     1 Scene \\
                                           MVS &                   \cite{jensen2014large} &           2014 &         \makecell[l]{Isolated Objects / \\ Focussed On \\ Objects} &                                               SCS &                                                                                     Stereo Camera &                                                                     Color, Depth &                                               - &                                                                  124 Scenes \\
 \rowcolor{gray!25}                              The Newer College &                 \cite{ramezani2020newer} &           2020 &                                                            Outdoor &          \makecell[l]{Structured Light, \\ LiDAR} &                                 \makecell[l]{Intel D435i, \\ Ouster OS-1 \\(Gen 1) 64} &                                                         Color, Depth, IMU &                                               - &                                                                    6 Scenes \\
                                      Megadepth &                     \cite{MegaDepthLi18} &           2018 &                                                            Outdoor &                                               SCS &                                                                                     Stereo Camera &                                                              Color, Depth &                                               - &                                                               130k Images \\
 \rowcolor{gray!25}     \makecell[l]{CVC-13: \\ Multimodal Stereo} &                           \cite{6220232} &           2013 &                                                            Outdoor &                                              SCS &                                                                                     Stereo Camera &                                                           Color, Infrared &                                               - &                                                                    4 Scenes \\
       \makecell[l]{Live Color+3D \\ Database} &                       \cite{su2013color} & 2011 &                                                            Outdoor &                                               TOF &                                           \makecell[l]{Range Scanner \\ (RIEGL VZ-400)} &                                                              Color, Depth &                                               - &                                                                   12 Scenes \\
 \rowcolor{gray!25}                                         Make3D &                           \cite{4531745} &           2009 &                                                            Outdoor &                                               TOF &                                               \makecell[l]{Custom-built 3-D \\ Scanner} &                                                              Color, Depth &                                               - &                                                                  534 Images \\
\makecell[l]{Fountain-P11 And \\ Herz-Jesu-P8} &                    \cite{Schoening2015a} &           2008 &                                                            Outdoor &                                             LiDAR &                                                                                      N/A &                                                              Color, Depth &                                               - &                                       \makecell[l]{2 Scenes (19 \\ Images)} \\
 \rowcolor{gray!25}                                             -- &                    \cite{beeler2011high} &           2011 &                            \makecell[l]{Partial Body W/o \\ Scene} &                                               SCS &                                                    \makecell[l]{Stereo Camera \\(seven Cameras)} &                                                                     Color, Depth &                                               - &                                                               N/A (2 Actors) \\
                                             -- &             \cite{QuattriniLiISERVO2016} &           2016 &                                                         Underwater &                                               SCS &                                                                                     Stereo Camera &                                 \makecell[l]{Color, IMU, \\ Sonar} &                                               - &                                                                 3 Sequences \\
 \rowcolor{gray!25}                                          DeMon &               \cite{ummenhofer2017demon} &           2017 &                                                                 N/A & \makecell[l]{SCS, Synthetic, \\ Structured Light} &                      \makecell[l]{Synthetic, Stereo\\ Camera, Asus Xtion \\ Pro Live, Kinect V1} &                                                              Color, Depth &                                               - &                                 \makecell[l]{20537 Sequences \\ And Scenes} \\
                                      Scenes11 &               \cite{ummenhofer2017demon} &           2017 &                                                                 N/A &                                                 - &                                                                               Synthetic &                                                              Color, Depth &                                               - &                                                             19959 Sequences \\

 \bottomrule[2pt]
\end{NiceTabular}

}
\end{table*}

\textbf{KITTI Dataset.} Analyzing the datasets presented in this paper, this is the most cited one. The KITTI Dataset consists of a complex system of IMU/GPS, LiDAR scanner, and multiple cameras ~\citep{Geiger2013IJRR}. They recorded 6 hours of traffic scenes and, in addition to collecting the information from the sensors, provided data from 3D object detection bounding boxes, optical flow, and visual odometry/SLAM~\citep{Geiger2012CVPR}. The project was expanded over the years, and the authors included data for tracking, road/lane detection, semantic/instance segmentation, and depth completion. Its depth completion data is composed of 94 thousand depth annotated RGB images~\citep{8374553} to produce dense depth maps from LiDAR points.

This dataset influenced the creation of the synthetic datasets Virtual KITTI~\citep{Gaidon:Virtual:CVPR2016} and Virtual KITTI 2~\citep{cabon2020vkitti2}. Recently, the KITTI authors released the KITTI-360 Dataset~\citep{Liao2021ARXIV}, which has more cameras, sensors, and more annotated data than the original KITTI Dataset.
\newline

\textbf{ScanNet Dataset.} ScanNet is an indoor dataset collected using an occipital structure sensor - a structured light sensor similar to Microsoft Kinect~v1~\citep{dai2017scannet}. The authors performed a dense reconstruction and conducted object instance-level annotation of all surfaces in the reconstruction. They also conducted a CAD Model Retrieval and Alignment for the objects in the scenes, which means that a 3D CAD model represented each instance of the annotated object in a scene. This dataset contains 2.5M views in 2,119 different scenes.
\newline

\textbf{SunCG Dataset.} The project associated with this dataset is focused on semantic scene completion, where from a single point of view, it estimates a complete 3D representation with the semantic label associated with the scene~\citep{song2016ssc}. Instead of estimating the semantic segmentation of visible surfaces, this project aims to predict the occluded space (3D scene representation) and a label for each voxel in the scene. Therefore, it deals with Reconstruction and Segmentation as a unified task. This dataset comprises synthetic data containing an entire 3D model scene (which can be related to reconstruction), with semantic labels associated with it.

\subsubsection{Segmentation or Other Extra Information}
 
In this sub-category, all datasets have extra information that leads to a better scene understanding. Extra information can be seen as semantic or instance segmentation, 2D or 3D object detection, optical flow, salient object detection, etc. For instance, datasets that explore potential applications for depth estimation algorithms and semantic segmentation, and datasets dedicated to salient object detection were categorized here.

The complete list of datasets containing extra information is available in Table~\ref{TabelaSOE}. We provide the type of extra information for each dataset in the ``Extra Data'' column. Researchers interested in a specific application, for instance, salient object detection, should use it to filter datasets related to their field of interest. Table~\ref{TabelaSOR_SOE} also reports datasets for this sub-category, as well as information of ``SLAM, Odometry, or Reconstruction'' sub-category. Therefore, researchers interested in semantic segmentation datasets may check both tables and refer to the ``Extra Data'' column to find the datasets that match their interest. Next, three of the most influencing and promising papers for this sub-category are presented.
\newline

\begin{table*}[!t]

\caption{Datasets of ``Segmentation or Other Extra Information''}
\label{TabelaSOE}
\resizebox{\textwidth}{!}{%

\begin{NiceTabular}{llllllllll}
\toprule[2pt]
                                                                           \textbf{Dataset Name} & \textbf{Ref.} & \textbf{Year} &     
\textbf{Scene Type} &  \textbf{Application} &   \textbf{Sensor Type} &                                            \textbf{Sensor Name} &                                                           \textbf{Data Modalities} &                                         \textbf{Extra Data} &                                                                                                          \textbf{Images/Scenes} \\
\midrule[1.5pt]

 \rowcolor{gray!25}
                                                VALID &                     \cite{9197186} & 2020 &                                                     Aerial &                                       SOE &                                         - &                                                                               Synthetic &                                                    Color, Depth & \makecell[l]{Object Detection, \\ Panoptic \\ Segmentation, \\ Instance \\ Segmentation, \\ Semantic \\ Segmentation} &                                                              \makecell[l]{6 Scenes (6690 \\ Images)} \\
                                                  US3D &                \cite{9frn-7208-20} & 2019 &                                                     Aerial &                                       SOE &                                     LiDAR &                                                                          Airborne LiDAR &                                                    Color, Depth &                                                                                \makecell[l]{Semantic \\ Segmentation} &               \makecell[l]{4160 Images From 3 \\ Different Cities \\ (a Fourth Is Not \\ Available)} \\
                                                   \rowcolor{gray!25}

                                       Vaihingen &                     \cite{haala2010german} &                                    2011 &          Aerial & SOE &                                   LiDAR &                                                                                       \makecell[l]{Leica ALS50 And \\ ALTM-ORION M} &                            Color, Depth &                                                                                             \makecell[l]{Semantic \\ Segmentation} &                                                                           33 Patches \\
                                                  
                                        Potsdam &                           \cite{rottensteiner2012isprs} &           2011 &                                                             Aerial & SOE &                                               N/A &                                                                                      N/A &                                                              Color, Depth &                                               \makecell[l]{Semantic \\ Segmentation} &                                                                  38 Patches \\

 \rowcolor{gray!25}                                        Leddar Pixset &            \cite{deziel2021pixset} & 2021 &                                                    Driving & \makecell[l]{SOE and Tracking \\ (Other)} &                                     LiDAR &                                                    \makecell[l]{Leddar Pixell \\ LiDAR} &                       \makecell[l]{Color, Depth, IMU, \\ Radar} &                                    \makecell[l]{3D Bounding Boxes, \\ 2D Bounding Boxes, \\ Semantic \\ Segmentation} &                                                           \makecell[l]{97 Sequences (29k \\ Frames)} \\
                                       Virtual Kitti 2 &            \cite{cabon2020vkitti2} & 2020 &                                                    Driving & \makecell[l]{SOE and Tracking \\ (Other)} &                                         - &                                                                               Synthetic &                                                    Color, Depth &                                  \makecell[l]{Semantic \\ Segmentation, \\ Instance \\ Segmentation, \\ Optical Flow} &                                    \makecell[l]{5 Scenes (multiple \\ Conditions For \\ Each Scene)} \\
 \rowcolor{gray!25}                                     Waymo Perception &          \cite{sun2020scalability} & 2020 &                                                    Driving &                                       SOE &                                     LiDAR &                                                                                      N/A &                                                    Color, Depth &                                                                                  \makecell[l]{3D Object \\ Detection} &                                                      \makecell[l]{1150 Scenes (20 \\ Seconds/scene)} \\
                                             Argoverse &                  \cite{ Argoverse} & 2019 &                                                    Driving & \makecell[l]{SOE and Tracking \\ (Other)} &                                SCS, LiDAR &                                                                         \makecell[l]{Argo LiDAR, \\Stereo Camera} &                                                    Color, Depth &                                                                                  \makecell[l]{3D Object \\ Detection} &                                                                                           113 Scenes \\
 \rowcolor{gray!25}                                           CityScapes &        \cite{Cordts2016Cityscapes} & 2016 &                                                    Driving &                                       SOE &                                       SCS &                                                                                     Stereo Camera &                                                 Color, Odometry &                                            \makecell[l]{Semantic \\ Segmentation, \\ 3D-object \\ Detection And Pose} &                                                            \makecell[l]{50 Cities (25k \\ Images)} \\
                                               SYNTHIA &               \cite{Ros_2016_CVPR} & 2016 &                                                    Driving &                                       SOE &  \makecell[l]{Virtual 8 Depth \\ Sensors} &                                                                               Synthetic &                                                    Color, Depth &                                                                                \makecell[l]{Instance \\ Segmentation} &            \makecell[l]{5 Sequences (with \\ Sub-sequences) At \\ 5 Fps. 200k Images \\ From Videos} \\
 \rowcolor{gray!25}          \makecell[l]{Daimler Urban \\ Segmentation} & \cite{scharwachter2014stixmantics} & 2014 &                                                    Driving &                                       SOE &                                       SCS &                                                                                     Stereo Camera &                                                           Color &                                                                                                     Semantic Labeling &                                                                                          5k Images \\
                  \makecell[l]{Ground Truth \\ Stixel} &      \cite{pfeiffer2013exploiting} & 2013 &                                                    Driving &                                       SOE &                                       SCS &                                                                                     Stereo Camera &                                                           Color &                                                                                                               Stixels &                                                                                         12 Sequences \\
 \rowcolor{gray!25}           \makecell[l]{Daimler Stereo \\ Pedestrian} &               \cite{keller2011new} & 2011 &                                                    Driving &                                       SOE &                                       SCS &                                                                                     Stereo Camera &                                                           Color &                                                                                                      Object Detection &                                                                                         28919 Images \\
                                                Unreal &                \cite{mancini2018j} & 2018 &                                           Driving, Outdoor &                                       SOE &                                         - &                                                                               Synthetic &                                                    Color, Depth &                                                                                \makecell[l]{Semantic \\ Segmentation} &                                                          \makecell[l]{21 Sequences (100k \\ Images)} \\
 \rowcolor{gray!25}                                             OASIS V2 &               \cite{chen2020oasis} & 2021 &                                                In-the-wild &                                       SOE &                                         - &                                                  \makecell[l]{From Human \\ Annotation} &                                                    Color, Depth &                                                                 \makecell[l]{Normal Maps, \\ Instance \\ Segmenation} &                                                                                        102k Images \\
                                                 OASIS &               \cite{chen2020oasis} & 2020 &                                                In-the-wild &                                       SOE &                                         - &                                                  \makecell[l]{From Human \\ Annotation} &                                                    Color, Depth &                                                                 \makecell[l]{Normal Maps, \\ Instance \\ Segmenation} &                                                                                        140k Images \\
 \rowcolor{gray!25}                                           RedWeb-S &                  \cite{9585702} & 2020 &                                                In-the-wild &                                       SOE &                          SCS &                                                                     Stereo Camera         &                                                    Color, Depth &                                                                                                         Saliency Mask &                                                                                          3179 Images \\

        DUTLF-Depth &                  \cite{Piao_2019_ICCV} & 2019 &                                                In-the-wild &                                       SOE &                          SCS &                                                                     \makecell[l]{Lytro Illum (Light \\ Field) (Stereo Camera)}         &                                                    Color, Depth &                                                                                                         Saliency Mask &                                                                                          1200 Images \\
                                                 
 \rowcolor{gray!25}                                           Scene Flow &                     \cite{MIFDB16} & 2016 &                                                In-the-wild &                                       SOE &                                         - &                                                                               Synthetic &                                                           Color &                                                                 \makecell[l]{Optical Flow, \\ Object \\ Segmentation} &                                                          \makecell[l]{2256 Scenes (39049 \\ Frames)} \\
 
         LFSD &                  \cite{li2014saliency} & 2015 &                                                In-the-wild &                                       SOE &                          SCS &                                                                     \makecell[l]{Lytro Illum (Light \\ Field) (Stereo Camera)}         &                                                    Color &                                                                                                         Saliency Mask &                                                                                          100 Images \\
 \rowcolor{gray!25}
        \makecell[l]{RGBD Salient \\ Object Detection} &                  \cite{PengECCV14} & 2014 &                                                In-the-wild &                                       SOE &                          Structured Light &                                                                               Kinect V1 &                                                    Color, Depth &                                                                                                         Saliency Mask &                                                                                          1000 Images \\
    MPI Sintel &            \cite{Butler:ECCV:2012} & 2012 &                                                In-the-wild &                                       SOE &                                         - &                                                                               Synthetic &                                                    Color, Depth &                                                                                                          Optical Flow &                                                         \makecell[l]{35 Scenes (50 \\ Frames/scene)} \\
  \rowcolor{gray!25} NYUv2-OC++ &      \cite{Ramamonjisoa_2020_CVPR} & 2020 &                                                     Indoor &                                       SOE &                          Structured Light &                                                                               Kinect V1 &                    \makecell[l]{Color, Depth, \\ Accelerometer} &                                                                            \makecell[l]{Occlusion \\ Boundaries Maps} &                                                              \makecell[l]{1449 Images From \\ NYUv2} \\
 \makecell[l]{Near-Collision \\ Set} &      \cite{manglik2019forecasting} & 2019 &                                                     Indoor &                                       SOE &                                SCS, LiDAR &                                                                         \makecell[l]{LiDAR (N/A),\\ Stereo Camera} &                                                    Color, Depth &                                                                                  \makecell[l]{2D Object \\ Detection} &                                                                                      13658 Sequences \\
 \rowcolor{gray!25}  SBM-RGBD  &                  \cite{camplani2017benchmarking} & 2017 &                                                Indoor &                                       SOE &                          Structured Light &                                                                     Kinect V1         &                                                    Color, Depth &                                                                                                         Saliency Mask &                                                                                          \makecell[l]{33 sequences \\ (~15000 frames)} \\
   SUN RGB-D &                 \cite{song2015sun} & 2015 &                                                     Indoor &                                       SOE & \makecell[l]{Structured Light \\ And TOF} & \makecell[l]{Intel RealSense 3D \\ Camera, Asus Xtion \\ LIVE PRO, Kinect \\V1 and V2} &                                                    Color, Depth &                                               \makecell[l]{Semantic \\ Segmentation, \\ Object Detection \\ And Pose} &                                                                                         10335 Images \\
 \rowcolor{gray!25} TUW &        \cite{aldoma2014automation} & 2014 &                                                     Indoor &                                       SOE &                          Structured Light &                                               \makecell[l]{ASUS Xtion \\ProLive RGB-D} &                                                    Color, Depth &                                                                          \makecell[l]{Object Instance \\ Recognition} &                                                           \makecell[l]{15 Sequences (163 \\ Frames)} \\
\makecell[l]{Willow And \\ Challenge} &        \cite{aldoma2014automation} & 2014 &                                                     Indoor &                                       SOE &                          Structured Light &                                                                               Kinect V1 &                                                    Color, Depth &                                                                          \makecell[l]{Object Instance \\ Recognition} &            \makecell[l]{24 Sequences (353 \\ Frames) For \\ Willow, 39 \\ Sequences (176 \\ Frames)} \\
 \rowcolor{gray!25} \makecell[l]{An In Depth View \\ of Saliency} &                  \cite{ciptadi-bmvc2013} & 2013 &                                                Indoor &                                       SOE &                          Structured Light &                                                                     Kinect V1         &                                                    Color, Depth &                                                                                                         Saliency Mask &                                                                                          80 Images \\
   NYU Depth V2 &            \cite{10.1007/978-3-642-33715-4_54} & 2012 &                                                     Indoor &                                       SOE &                          Structured Light &                                                                               Kinect V1 &                    \makecell[l]{Color, Depth, \\ Accelerometer} &                                                                                \makecell[l]{Semantic \\ Segmentation} &             \makecell[l]{464 Scenes (407024 \\ Frames) With 1449 \\ Labeled Aligned \\ RGB-D Images} \\

 \bottomrule[2pt]
\end{NiceTabular}

}

\vspace{-0.3cm}
 \begin{flushright}\scriptsize Continue on Next Page \end{flushright}
\end{table*}

\begin{table*}[!t]

\resizebox{\textwidth}{!}{%

\begin{NiceTabular}{llllllllll}
\toprule[2pt]
                                                                           \textbf{Dataset Name} & \textbf{Ref.} & \textbf{Year} &     
\textbf{Scene Type} &  \textbf{Application} &   \textbf{Sensor Type} &                                            \textbf{Sensor Name} &                                                           \textbf{Data Modalities} &                                         \textbf{Extra Data} &                                                                                                          \textbf{Images/Scenes} \\
\midrule[1.5pt]

 \rowcolor{gray!25}   --    &             \cite{mason2012object} & 2012 &                                                     Indoor &                                       SOE &                          Structured Light &                                                                               Kinect V1 &                                                    Color, Depth &                                                                                \makecell[l]{Semantic \\ Segmentation} &                                       \makecell[l]{3 Options. Large: 2 \\ Sequences (397 \\ Frames)} \\

  Berkeley B3DO &          \cite{janoch2013category} & 2011 &                                                     Indoor &                                       SOE &                          Structured Light &                                                                               Kinect V1 &                                                    Color, Depth &                                                                                                      Object Detection &                                                              \makecell[l]{75 Scenes (849 \\ Images)} \\

 \rowcolor{gray!25} NYU Depth V1 &           \cite{silberman11indoor} & 2011 &                                                     Indoor &                                       SOE &                          Structured Light &                                                                               Kinect V1 &                                                    Color, Depth &                                                                                \makecell[l]{Semantic \\ Segmentation} &                      \makecell[l]{64 Scenes (108617 \\ Frames) With 2347 \\ Labeled RGB-D \\ Frames} \\
  COTS  &                  \cite{9340352} & 2021 &                                                \makecell[l]{Isolated Objects / \\ Focussed On \\ Objects} &                                       SOE &                          SCS &                                                                     \makecell[l]{Intel Realsense \\ D435}        &                                                    Color, Depth &                                                                                                         Saliency Mask &                                                                                          120 Images \\
 \rowcolor{gray!25} ClearGrasp &             \cite{sajjan2020clear} & 2019 & \makecell[l]{Isolated Objects / \\ Focussed On \\ Objects} &                                       SOE &                          SCS &                                        \makecell[l]{Synthetic, Intel \\ RealSense D415} &                                                    Color, Depth &                                                 \makecell[l]{Normal Maps, \\ Semantic \\ Segmentation - \\ Synthetic} &    \makecell[l]{Over 50k \\ Synthetic Images \\ Of 9 Objects. 286 \\ Real Images Of 10 \\ Objects} \\
  T-LESS &              \cite{hodan2017tless} & 2017 & \makecell[l]{Isolated Objects / \\ Focussed On \\ Objects} &                                       SOE &    \makecell[l]{Structured Light, \\ TOF} &                                   \makecell[l]{Kinect V2, \\ PrimeSense \\ Carmine 1.0} &                                                    Color, Depth &                                                                             \makecell[l]{3D Instance \\ Segmentation} & \makecell[l]{N/A Scenes (38k \\ Images) For \\ Training. 20 \\ Scenes (10k \\ Images) For \\ Testing} \\
 \rowcolor{gray!25} DROT &             \cite{rotman2016depth} & 2016 & \makecell[l]{Isolated Objects / \\ Focussed On \\ Objects} &                                       SOE &    \makecell[l]{Structured Light, \\ TOF} &                                       \makecell[l]{Kinect V1, V2 And \\ RealSense R200} &                                                    Color, Depth &                                                                                                         Object Motion &                                                               \makecell[l]{5 Scenes (112 \\ Frames)} \\
 MPII Multi-Kinect &               \cite{susanto20123d} & 2012 & \makecell[l]{Isolated Objects / \\ Focussed On \\ Objects} &                                       SOE &    \makecell[l]{SCS, Structured \\ Light} &                                                                         \makecell[l]{ Kinect V1, \\Stereo Camera} &                                                    Color, Depth &                                                                                                      Object Detection &                                                              \makecell[l]{33 Scenes (560 \\ Images)} \\
 \rowcolor{gray!25}   Mid-Air &            \cite{Fonder2019MidAir} & 2019 &                                                    Outdoor &                                       SOE &                                         - &                                                                               Synthetic & \makecell[l]{Color, Depth, \\ Accelerometer, \\ Gyroscope, GPS} &                                                                \makecell[l]{Normal Maps, \\ Semantic \\ Segmentation} &                                                       \makecell[l]{54 Sequences \\ (420k Frames)} \\

 \bottomrule[2pt]

  \multicolumn{7}{l}{SOE: Segmentation or Other Extra Information} 
 
\end{NiceTabular}

}

\end{table*}

\begin{table*}[!t]

\caption{Datasets of ``SLAM, Odometry, or Reconstruction'' and ``Segmentation or Other Extra Information' Categories''}
\label{TabelaSOR_SOE}
\resizebox{\textwidth}{!}{%

\begin{NiceTabular}{llllllllll}
\toprule[2pt]
                                                                           \textbf{Dataset Name} & \textbf{Ref.} & \textbf{Year} &     
\textbf{Scene Type} &  \textbf{Application} &   \textbf{Sensor Type} &                                            \textbf{Sensor Name} &                                                           \textbf{Data Modalities} &                                         \textbf{Extra Data} &                                                                                                          \textbf{Images/Scenes} \\
\midrule[1.5pt]

 \rowcolor{gray!25}
                                          -- &                          \cite{9412256} &           2020 &                                                             Aerial &                                              SOR and SOE &                                        - &                                                                             Synthetic &                                                                         Color, Depth &                                                                                                                             \makecell[l]{Normal Maps, \\ Edges, Semantic \\ Labels} &                                                \makecell[l]{15 Scenes (144k \\ Images)} \\
                                   EventScape &               \cite{RAL21Gehrig} &           2021 &                                                            Driving &                                              SOR and SOE &                                        - &                                                                             Synthetic &                                                                         Color, Depth &                                                               \makecell[l]{Semantic \\ Segmentation, \\ Navigation Data \\ (Position, \\ Orientation, \\ Angular Velocity, \\ Etc)} &                                                                             758 Sequences \\
 \rowcolor{gray!25}                                   KITTI-360 &             \cite{Liao2021ARXIV} &           2021 &                                                            Driving &                                              SOR and SOE &                               SCS, LiDAR &                                   \makecell[l]{Velodyne (LiDAR) \\ Points Cloud, \\ Stereo Camera} &                                              \makecell[l]{Color, Depth, GPS, \\ IMU} & \makecell[l]{2D-object \\ Detection, \\ 3D-object \\ Detection, \\ Tracking, \\ Instance \\ Segmentation, \\ Optical Flow. \\ These Are Not In \\ Necessary In The \\ Same Dataset} &              \makecell[l]{11 Sequences To \\ Over 320k Images \\ And 100k Laser \\ Scans} \\
                                         DDAD &                   \cite{packnet} &      2020 &                                                            Driving &                                              SOR and SOE &                                    LiDAR &                                                                      Luminar-H2 LiDAR &                                                                          Color, Deph &                                                                                                                                              \makecell[l]{Instance \\ Segmentation} &                                                \makecell[l]{150 Scenes (12650 \\ Frames)} \\
 \rowcolor{gray!25}                                Lyft Level 5 &     \cite{houston2020one} &           2020 &                                                            Driving &                                              SOR and SOE &                               SCS, LiDAR &                    \makecell[l]{3 LiDAR (40 And \\ 64-beam LiDARs), 5 \\ Radars, Stereo Camera} &                                                 \makecell[l]{Color, Depth, \\ Radar} &                                                                                                                                                \makecell[l]{3D Object \\ Detection} &                                          \makecell[l]{170k Scenes (25 \\ Seconds Each)} \\
 NuScenes &              \cite{nuscenes2019} &      2020 &                                                            Driving &                                              SOR and SOE &                                    LiDAR &                                                                                    N/A &                                            \makecell[l]{Color, Depth, \\ Radar, IMU} &                                                                                                                   \makecell[l]{3D Object \\ Detection, \\ Semantic \\ Segmentation} &   \makecell[l]{1000 Scenes (20 \\ Seconds Each). \\ 1.4M Images And \\ 390k LiDAR Sweeps} \\
 \rowcolor{gray!25}                 
                                     Woodscape &     \cite{yogamani2019woodscape} & 2019 &                                                            Driving &                                              SOR and SOE &                                    LiDAR &                                                                      Velodyne HDL-64E &                                              \makecell[l]{Color, IMU, GPS, \\ Depth} &                                                                                                                      \makecell[l]{Instance \\ Segmentation, 2D \\ Object Detection} &                                               \makecell[l]{50 Sequences (100k \\ Frames)} \\
                                Virtual Kitti &   \cite{Gaidon:Virtual:CVPR2016} &           2016 &                                                            Driving &                                              SOR and SOE &                                        - &                                                                             Synthetic &                                                                         Color, Depth &                                                                                                \makecell[l]{Semantic \\ Segmentation, \\ Instance \\ Segmentation, \\ Optical Flow} &                                                 \makecell[l]{50 Videos (21260 \\ Frames)} \\
 \rowcolor{gray!25}                                       KITTI &            \cite{Geiger2013IJRR} &           2012 &                                                            Driving &                                              SOR and SOE &                               SCS, LiDAR &                                   \makecell[l]{Velodyne (LiDAR) \\ Points Cloud,\\ Stereo Camera} &                                   \makecell[l]{Color, Grayscale, \\ Depth, GPS, IMU} &                                                                                                                                              \makecell[l]{Instance \\ Segmentation} &                                                 \makecell[l]{61 Scenes (42746 \\ Frames)} \\
                                     Hypersim &              \cite{roberts:2021} &           2021 &                                                             Indoor &                                              SOR and SOE &                                        - &                                                                             Synthetic &                                                                         Color, Depth &                                                                                                   \makecell[l]{Normal Maps, \\ Instance \\ Segmentation, \\ Diffuse \\ Reflectance} &                                                \makecell[l]{461 Scenes (77400 \\ Images)} \\
 \rowcolor{gray!25}                                    RoboTHOR &        \cite{deitke2020robothor} &           2020 &                                                             Indoor &                                              SOR and SOE &                                        - &                                                                             Synthetic &                                                                         Color, Depth &                                                                                                                                              \makecell[l]{Instance \\ Segmentation} &                                                                                 75 Scenes \\
                                 Structured3D &     \cite{zheng2020structured3d} &           2020 &                                                             Indoor &                                              SOR and SOE &                                        - &                                                                             Synthetic &                                                                         Color, Depth &                                                                                                                         \makecell[l]{Object Detection, \\ Semantic \\ Segmentation} &                          \makecell[l]{3500 Scenes With \\ 21835 Rooms \\ (196515 Frames)} \\
 \rowcolor{gray!25}                                     Replica &            \cite{replica19arxiv} &           2019 &                                                             Indoor &                                              SOR and SOE &                         Structured Light &                                                                                    N/A &                                 \makecell[l]{Color, Depth, IMU, \\ Grayscale Camera} &                                                                                                                              \makecell[l]{Normal Maps, \\ Instance \\ Segmentation} &                                                                                 18 Scenes \\
                                       Gibson &   \cite{xiazamirhe2018gibsonenv} &           2018 &                                                             Indoor &                                              SOR and SOE & \makecell[l]{LiDAR, Structured \\ Light} &                          \makecell[l]{NavVis, \\ Matterport \\ Camera, \\ DotProduct} &                                                                         Color, Depth &                                                                                                                              \makecell[l]{Normal Maps, \\ Semantic \\ Segmentation} &                      \makecell[l]{572 Scenes. 1400 \\ Floor Spaces From \\ 572 Buildings} \\
 \rowcolor{gray!25}                                 InteriorNet &            \cite{ InteriorNet18} &           2018 &                                                             Indoor &                                              SOR and SOE &                                        - &                                                                             Synthetic &                                                                    Color, Depth, IMU &                                                                                                                              \makecell[l]{Normal Maps, \\ Semantic \\ Segmentation} &                                                                         20 Million Images \\
                                    Taskonomy &             \cite{8578148} &           2018 &                                                             Indoor &                                              SOR and SOE &                         Structured Light &                                                                                    N/A &                                                                         Color, Depth &                                                                              \makecell[l]{25 Tags (Normals \\ Maps, Semantic \\ Segmentation, \\ Scene \\ Classification, \\ Etc.)} &                                                                        4.5 Million Scenes \\
 \rowcolor{gray!25}                                         AVD & \cite{active-vision-dataset2017} &           2017 &                                                             Indoor &                                              SOR and SOE &                                      TOF &                                                                             Kinect V2 &                                                                         Color, Depth &                                                                                                                                                                    Object Detection &                                              \makecell[l]{15 Scenes (over 30k \\ Images)} \\
                                 MatterPort3D &              \cite{Matterport3D} &           2017 &                                                             Indoor &                                              SOR and SOE &   \makecell[l]{SCS, Structured \\ Light} &                                               \makecell[l]{Matterport \\ Camera, Stereo\\ Camera} &                                                                         Color, Depth &                                                                                                        \makecell[l]{Semantic \\ Segmentation, 3D \\ Semantic-voxel \\ Segmentation} &                      \makecell[l]{90 Scenes, 10800 \\ Panoramic Views \\ (194400 Images)} \\
 \rowcolor{gray!25}                                     ScanNet &            \cite{dai2017scannet} &           2017 &                                                             Indoor &                                              SOR and SOE &                         Structured Light & \makecell[l]{Occipital \\ Structure Sensor - \\ Similar to \\ Kinect V1} &                                                                         Color, Depth &                                                                                                                                     \makecell[l]{3D Semantic-voxel \\ Segmentation} &                              \makecell[l]{1513 Sequences \\ (over 2.5 Million \\ Frames)} \\
                               SceneNet RGB-D &    \cite{McCormac:etal:ICCV2017} &           2017 &                                                             Indoor &                                              SOR and SOE &                                        - &                                                                             Synthetic &                                                                         Color, Depth &                                                                                                                             \makecell[l]{Instance \\ Segmentation, \\ Optical Flow} &                                 \makecell[l]{15K Trajectories \\ (scenes) (5M \\ Images)} \\
 \rowcolor{gray!25}                                       SunCG &               \cite{song2016ssc} &           2017 &                                                             Indoor &                                              SOR and SOE &                                        - &                                                                             Synthetic &                                                                         Color, Depth &                                                                                                                                              \makecell[l]{Semantic \\ Segmentation} &                                                                              45622 Scenes \\
 
                                   GMU Kitchen &    \cite{georgakis2016multiview} &           2016 &                                                             Indoor &                                              SOR and SOE &                                      TOF &                                                                             Kinect V2 &                                                                         Color, Depth &                                                                                                                                                                    Object Detection &                                                   \makecell[l]{9 Scenes (6735 \\ Frames)} \\
 
 \bottomrule[2pt]
\end{NiceTabular}

}
\vspace{-0.3cm}
 \begin{flushright}\scriptsize Continue on Next Page \end{flushright}

\end{table*}

\begin{table*}[!t]

\resizebox{\textwidth}{!}{%

\begin{NiceTabular}{llllllllll}
\toprule[2pt]
                                                                           \textbf{Dataset Name} & \textbf{Ref.} & \textbf{Year} &     
\textbf{Scene Type} &  \textbf{Application} &   \textbf{Sensor Type} &                                            \textbf{Sensor Name} &                                                           \textbf{Data Modalities} &                                         \textbf{Extra Data} &                                                                                                          \textbf{Images/Scenes} \\
\midrule[1.5pt]

\rowcolor{gray!25}                                Stanford2D3D &       \cite{2017arXiv170201105A} &           2016 &                                                             Indoor &                                              SOR and SOE &                         Structured Light &                                                                     Matterport Camera &                                                                         Color, Depth &                                                                                                                              \makecell[l]{Semantic \\ Segmentation, \\ Normal Maps} &                             \makecell[l]{6 Large-scale \\ Indoor Areas \\ (70496 Images)} \\

                                        SUN3D &             \cite{xiao2013sun3d} &           2013 &                                                             Indoor &                                              SOR and SOE &                         Structured Light &                                                                   \makecell[l]{Asus Xtion \\ Pro Live} &                                                                         Color, Depth &                                                                                                                                              \makecell[l]{Semantic \\ Segmentation} &                                                                             415 Sequences \\

 \rowcolor{gray!25}                                     Starter &      \cite{eftekhar2021omnidata} &           2021 &                               \makecell[l]{Indoor, \\ In-the-wild} & \makecell[l]{SOR And SOE \\ (depending On \\ Subdataset)} &                         Structured Light &                                    \makecell[l]{Synthetic, \\ Matterport Pro2, \\ NA} & \makecell[l]{Color, Depth, IMU, \\ Grayscale Camera \\ (depending On \\ Subdataset)} &                                                           \makecell[l]{Normals Maps, \\ Semantic \\ Segmentation, \\ Scene \\ Classification, \\ Etc. (depending On \\ Subdataset)} &                                      \makecell[l]{Over 14.6M Images \\ (multiple Scenes)} \\
                                  RGBD Object &              \cite{lai2011large} &           2011 & \makecell[l]{Indoor, Isolated \\ Objects / Focussed \\ On Objects} &                                              SOR and SOE &                         Structured Light &                                                                             Kinect V1 &                                                                         Color, Depth &                                                                                                                                                                     3D Segmentation &                  \makecell[l]{8 Sequences And 300 \\ Isolated Objects \\ (250k Frames)} \\
 \rowcolor{gray!25}                                   TartanAir &         \cite{tartanair2020iros} &           2020 &                                                    Indoor, Outdoor &                                              SOR and SOE &                                        - &                                                                       Synthetic LiDAR &                                                                         Color, Depth &                                                                                                                             \makecell[l]{Semantic \\ Segmentation, \\ Optical Flow} & \makecell[l]{1037 Scenes (Over \\ 1M Frames). Each \\ Scene Contains \\ 500-4000 Frames.} \\
 \makecell[l]{RGB-D Semantic \\ Segmentation} &         \cite{tombari2011online} &           2011 &         \makecell[l]{Isolated Objects / \\ Focussed On \\ Objects} &                                              SOR and SOE &                         Structured Light &                                                                             Kinect V1 &                                                                         Color, Depth &                                                                                                                                           \makecell[l]{3D Semantic \\ Segmentation} &                                                                            16 Test Scenes \\
 \rowcolor{gray!25}                                     GTA-SfM &              \cite{wang2020flow} &           2020 &                                                            Outdoor &                                              SOR and SOE &                                        - &                                                                             Synthetic &                                                                         Color, Depth &                                                                                                                                                                        Optical Flow &                                                                              76k Images \\

 \bottomrule[2pt]
 
 \multicolumn{7}{l}{SOR: SLAM, Odometry, or Reconstruction} \\
  \multicolumn{7}{l}{SOE: Segmentation or Other Extra Information}

\end{NiceTabular}

}
\end{table*}

\textbf{NYUv2.} This dataset contains indoor images and is the most cited dataset for this type of scene in the ``Segmentation or Other Extra Information'' sub-category. It was collected using Microsoft Kinect~v1 sensor and is composed of aligned RGB and depth images, labeled data containing semantic segmentation, and raw data~\citep{10.1007/978-3-642-33715-4_54}. This project is a continuation of NYUv1~\citep{silberman11indoor}, which uses the same sensor and type of data, but has fewer scenes and total frames. 
\newline

\textbf{Scene Flow Datasets.} This dataset is a collection of three datasets: FlyingThing3D, Monkaa, and Driving. The first is composed of everyday objects flying along random trajectories~\citep{MIFDB16}. The second was created using Blender computer graphics software, based on the information from an animated short film called Monkaa. The third is composed of a street scene. Scene Flow contains only synthetic data for all three datasets and, in addition to depth and RGB frames, the authors also include optical flow, segmentation, and stereo disparity change data.
\newline

\textbf{Waymo Perception.} This dataset is a street scene dataset composed of RGB and LiDAR labels. It consists of street scenes, and the authors labeled LiDAR using 3D bounding boxes for vehicles, pedestrians, cyclists, and signs~\citep{sun2020scalability}. They also provide RGB images annotations with 2D bounding boxes of vehicles, pedestrians, and cyclists. The 3D bounding boxes also have unique tracking IDs for tracking applications. The Waymo Perception Dataset is composed of 1,150 scenes with 20 seconds of recording each.

\subsubsection{Depth Data Only}

\begin{table*}[!t]

\caption{Datasets of ``Depth Data Only'' sub-category}
\label{TabelaDepth}
\resizebox{\textwidth}{!}{%

\begin{NiceTabular}{lllllllll}
\toprule[2pt]
                                                                           \textbf{Dataset Name} & \textbf{Ref.} & \textbf{Year} &     
\textbf{Scene Type} &  \textbf{Sensor Type} &                                              \textbf{Sensor Name} &                                                           \textbf{Data Modalities} &                                         \textbf{Extra Data} &                                                                                                          \textbf{Images/Scenes} \\
\midrule[1.5pt]

 \rowcolor{gray!25}
                                         Espada &                     \cite{9507349} &                                    2021 &          Aerial &                                        - &                                                                                                                           Synthetic &                            Color, Depth &                                                                                             - &                                        \makecell[l]{49 Environments \\ (80k Images)} \\
                                            DSEC &                 \cite{Gehrig21ral} &                                    2021 &         Driving &                               SCS, LiDAR &                                                                                               \makecell[l]{Velodyne VLP-16, \\ Stereo Camera} &                       Color, Depth, GPS &                                                                                             - &                                                                         53 Sequences \\
\rowcolor{gray!25}                                       Mapillary &      \cite{antequera2020mapillary} &                                    2020 &         Driving &                                      SCS &                                                                                                                                 Stereo Camera &                            Color, Depth &                                                                                             - &                                        \makecell[l]{50k Scenes \\ (750k Images)} \\
            \makecell[l]{RabbitAI \\ Benchmark} &           \cite{schilling2020mind} &                                    2020 &         Driving &                                 SCS &                                                                                        \makecell[l]{17-camera \\ Light-field} &                                   Color, Depth &                                                                                             - &                   \makecell[l]{200 Scenes (100 For \\ Training, 100 For \\ Testing)} \\
\rowcolor{gray!25} \makecell[l]{DrivingStereo - \\ Driving Stereo} &       \cite{yang2019drivingstereo} &                                    2019 &         Driving &                               SCS, LiDAR &                                                                                              \makecell[l]{Velodyne HDL-64E, \\ Stereo Camera} & \makecell[l]{Color, Depth, IMU, \\ GPS} &                                                                                             - &                                        \makecell[l]{42 Sequences \\ (182188 Frames)} \\
           \makecell[l]{Urban Virtual \\ (UVD)} &           \cite{mancini2017toward} &                                    2017 &         Driving &                                        - &                                                                                                                           Synthetic &                            Color, Depth &                                                                                             - &                                                                         58500 Images \\
\rowcolor{gray!25}                                    DiverseDepth &         \cite{yin2020diversedepth} &                                    2020 &     In-the-wild &                                      SCS &                                                                                                                                 Stereo Camera &                                   Color &                                                                                             - &                                                                        320k Images \\
                                           HRWSI &              \cite{Xian_2020_CVPR} &                                    2020 &     In-the-wild &                                      SCS &                                                                                                                                 Stereo Camera &                            Color, Depth &                                                                                             - &                                                                         20778 Images \\
\rowcolor{gray!25}                                      Holopix50k &           \cite{hua2020holopix50k} &                                    2020 &     In-the-wild &                                      SCS &                                                                                                                                 Stereo Camera &                                   Color &                                                                                             - &                                                                         49368 Images \\
                                     DualPixels &      \cite{GargDualPixelsICCV2019} &                                    2019 &     In-the-wild &                                      SCS &                                                                                                                                 Stereo Camera &                            Color, Depth &                                                                                             - &                                                                          3190 Images \\
\rowcolor{gray!25}                                       TAU Agent &              \cite{gil2019monster} &                                    2019 &     In-the-wild &                                        - &                                                                                                                           Synthetic &                            Color, Depth &                                                                                             - &                                                                             5 Scenes \\
                                           WSVD &                 \cite{wang2019web} &                                    2019 &     In-the-wild &                                      SCS &                                                                                                                                 Stereo Camera &                            Color, Depth &                                                                                             - &                                         \makecell[l]{553 Videos \\ (1.5M Frames)} \\
\rowcolor{gray!25}                                          ReDWeb &              \cite{Xian_2018_CVPR} &                                    2018 &     In-the-wild &                                      SCS &                                                                                                                                 Stereo Camera &                            Color, Depth &                                                                                             - &                                                                          3600 Images \\
             \makecell[l]{AirSim \\ Building\textunderscore99} &                 \cite{KIBWFC_2021} &                                    2021 &          Indoor &                                        - &                                                                                                                           Synthetic &                            Color, Depth &                                                                                             - &                                                                         20k Images \\
\rowcolor{gray!25}                                          Pano3D &           \cite{albanis2021pano3d} &                                    2021 &          Indoor & \makecell[l]{LiDAR, Structured \\ Light} & \makecell[l]{3 RGB Cameras. The \\3 Depth Cameras. \\ Matterport \\ Camera, NavVis, \\ DotProduct \\ (depending On \\ Subdataset)} &                            Color, Depth &                                                                                   Normal Maps &                                                                        42923 Samples \\
             \makecell[l]{Multiscopic \\ Vision} &            \cite{yuan2020mfusenet} &                                    2020 &          Indoor &                                      - &                                                                                                                       Synthetic &                                   Color &                                                                                             - & \makecell[l]{Around 1200 Scenes \\ Of Synthetic Data, \\ 92 Scenes Of Real \\ Data.} \\
\rowcolor{gray!25}                                             IRS &                 \cite{wang2019irs} &                               2019 &          Indoor &                                        - &                                                                                                                           Synthetic &                            Color, Depth &                                                                                   Normal Maps &                                                                        100025 Images \\
                                        IBims-1 & \cite{ 10.1016/j.cviu.2019.102877} &                               2019 &          Indoor &                                      TOF &                                                                             \makecell[l]{Laser (Leica \\ HDS7000 Laser \\ Scanner)} &                            Color, Depth & \makecell[l]{Semantic \\ Segmentation \\ (only For Planar \\ Areas: Walls, \\ Tables, Floor)} &                                                                           100 Images \\
\rowcolor{gray!25}                                 Middlebury 2014 &          \cite{scharstein2014high} &                                    2014 &          Indoor &                                      SCS &                                                                                                                                 Stereo Camera &                            Color, Depth &                                                                                             - &                                                                            33 Images \\
                                Middlebury 2006 &      \cite{scharstein2007learning} &                                    2006 &          Indoor &                         Structured Light &                                                                                      \makecell[l]{Custom-build \\ Structured Light} &                            Color, Depth &                                                                                             - &                                                                            21 Images \\
\rowcolor{gray!25}                                 Middlebury 2005 &      \cite{scharstein2007learning} &                                    2005 &          Indoor &                         Structured Light &                                                                                      \makecell[l]{Custom-build \\ Structured Light} &                            Color, Depth &                                                                                             - &                                                                             9 Images \\
                                Middlebury 2003 &          \cite{scharstein2003high} &                                    2003 &          Indoor &                         Structured Light &                                                                                      \makecell[l]{Custom-build \\ Structured Light} &                            Color, Depth &                                                                                             - &                                                                             2 Images \\
\rowcolor{gray!25}                                 Middlebury 2001 &          \cite{scharstein2003high} &                                    2001 &          Indoor &                                      SCS &                                                                                                                                 Stereo Camera &                            Color, Depth &                                                                                             - &                                                                             6 Images \\
                                          DIODE &               \cite{diode_dataset} &                                    2019 & Indoor, Outdoor &                                    LiDAR &                                                                                                                     FARO Focus S350 &                            Color, Depth &                                                                                   Normal Maps &          \makecell[l]{30 Scenes (8574 \\ Indoor Images, \\ 16884 Outdoor \\ Images)} \\
\rowcolor{gray!25}                                        DIML/CVL &                 \cite{cho2021deep} & \makecell[l]{2016, 2017, \\2018, 2021} & Indoor, Outdoor &                                 SCS, TOF &                                                     \makecell[l]{Kinectv2 For \\ Indoor, ZED Stereo \\ Camera \\  for Outdoor} &                            Color, Depth &                                                                                             - &                                                \makecell[l]{More Than 200 \\ Scenes} \\
          \makecell[l]{Forest Virtual \\ (FVD)} &           \cite{mancini2017toward} &                                    2017 &         Outdoor &                                        - &                                                                                                                           Synthetic &                            Color, Depth &                                                                                             - &                                                                         49500 Images \\
\rowcolor{gray!25}                                   Zurich Forest &           \cite{mancini2017toward} &                                    2017 &         Outdoor &                                      SCS &                                                                                                                                 Stereo Camera &                            Color, Depth &                                                                                             - &                                           \makecell[l]{3 Sequences (9846 \\ Images)} \\
                                             -- &           \cite{cui2021underwater} &                                    2021 &      Underwater &                                      SCS &                                                                                                                                 Stereo Camera &                            Color, Depth &                                                                                             - &                           \makecell[l]{600 Pairs (51 With \\ Depth Ground \\ Truth)} \\
\rowcolor{gray!25}                                           SQUID &        \cite{berman2020underwater} &                                    2020 &      Underwater &                                      SCS &                                                                                                                                 Stereo Camera &                            Color, Depth &                                                                                             - &                                                                            57 Images \\
 \bottomrule[2pt]
\end{NiceTabular}

}
\end{table*}

The datasets presented here are for the specific purpose of training depth estimation algorithms. They do not directly  provide reconstruction, SLAM, or other information, although some of these applications are direct results of depth estimation. For example, these works explore monocular depth estimation~\citep{CHO2021114877}, zero-shot depth estimation~\citep{yin2020diversedepth}, and multi-camera depth estimation~\citep{antequera2020mapillary}.

We present all papers found specifically for depth estimation in Table~\ref{TabelaDepth}. All datasets for all categories and sub-categories in this paper also contain depth information as it is an inclusion criterion for papers to be incorporated to this work. Some relevant papers in this sub-category are:
\newline

\textbf{ReDWeb Dataset.} This dataset deals with the in-the-wild scenario, covering scenes such as street, office, park, farm, etc. As formed in the acronym of this dataset's name ``Relative Depth from Web'' (ReDWeb), this dataset is formed by stereo images collected from the Internet~\citep{Xian_2018_CVPR}. The authors use optical flow to generate correspondence maps and create a relative depth map of the image. They post-process the data by segmenting the sky to increase the quality of the depth maps.
\newline

\textbf{SQUID Dataset.} This dataset is composed of underwater images collected from four different sites: two in the Red Sea and two in the Mediterranean Sea~\citep{berman2020underwater}. In addition to collecting stereo pair images, the authors included a ColorChecker to propose color restoration techniques in underwater images. 
\newline

\textbf{Middlebury Datasets.} These datasets are a composition of data released in different papers over the years of 2001, 2003, 2005, 2006, and 2014. These datasets are acquired using different strategies: custom structured light using a video projector for the Middlebury 2003~\citep{scharstein2003high}, Middlebury 2005~\citep{scharstein2007learning, Hirschmller2007EvaluationOC}, Middlebury 2006~\citep{scharstein2007learning, Hirschmller2007EvaluationOC}, and Middlebury 2014~\citep{scharstein2014high}, while Middlebury 2001~\citep{scharstein2002taxonomy} uses stereo image pair disparities. Despite using a custom structure light system, Middlebury 2014 contains improvements in the acquisition process.

\subsubsection{Other}

This sub-category contains all datasets that do not fit into the previous divisions. There is no sub-category in ``Other'' with more than four examples. Therefore, we did not create a specific sub-section for them.

All datasets here contain depth data and are divided into the following applications: novel view synthesis, foggy images for visibility restoration, relative depth between pairs of random points, object tracking, depth refinement for mirror surfaces, and synthesis of 4D RGB-D light field images. In Table~\ref{Tabela9}, we display all these datasets and their respective application as a column of the table. The most cited dataset included here is: 
\newline

\textbf{FRIDA2.} This dataset is a synthetic dataset of foggy images of the street view. It is formed by 330 synthetic images of 66 different scenes, where each image without fog is associated with four images that vary the intensity of the artificial fog presented in it~\citep{tarel2012vision}. Therefore, 66 images without fog have one depth map and four foggy images associated with it. FRIDA2 is a continuation of The Foggy Road Image DAtabase (FRIDA)~\citep{tarel2010improved}, which has similar characteristics to FRIDA2, but fewer images (only 18 distinct scenes). These datasets are created for image enhancement in foggy images, trying to reduce the impact of the fog in the visibility of street scenes.
\newline


\begin{table*}[!t]

\caption{``Other'' Scene/Objects RGB-D Datasets}
\label{Tabela9}
\resizebox{\textwidth}{!}{%

\begin{NiceTabular}{llllllllll}
\toprule[2pt]
                                                                           \textbf{Dataset Name} & \textbf{Ref.} &\textbf{Year} &     
\textbf{Scene Type} & \textbf{Application} &  \textbf{Sensor Type} &                                              \textbf{Sensor Name} &                                                           \textbf{Data Modalities} &                                         \textbf{Extra Data} &                                                                                                          \textbf{Images/Scenes} \\
\midrule[1.5pt]

 \rowcolor{gray!25}
                                          FRIDA2 &        \cite{tarel2012vision} & 2012 &         Driving &                                       Fog (Other) &                                    - &                                                                                                                         Synthetic &                                    Color, Depth &                                      - &  \makecell[l]{66 Scenes (330 \\ Images)} \\
                                            FRIDA &      \cite{tarel2010improved} & 2010 &         Driving &                                       Fog (Other) &                                    - &                                                                                                                         Synthetic &                                    Color, Depth &                                      - &   \makecell[l]{18 Scenes (90 \\ Images)} \\
 \rowcolor{gray!25}                                    3D Ken Burns &       \cite{Niklaus_TOG_2019} & 2019 &     In-the-wild &             \makecell[l]{3D Ken Burns \\ (Other)} &                                    - &                                                                                                                         Synthetic &                                    Color, Depth &                            Normal Maps &                             46 Sequences \\
                                              DIW &      \cite{NIPS2016_0deb1c54} & 2016 &     In-the-wild &                                    Points (Other) &                                    - &                                                                                \makecell[l]{Two Points \\ (manually \\ Anotated)} & \makecell[l]{Color, Depth \\ Points (2 Points)} &                                      - &                            495k Images \\
 \rowcolor{gray!25}                                        Mirror3D &        \cite{mirror3d2021tan} & 2021 &          Indoor &                                    Mirror (Other) & \makecell[l]{Structured \\ Light, SCS} & \makecell[l]{Matterport Camera, \\ Stereo Camera, \\ Kinect V1, \\ Occipital \\ Structure Sensor - \\ Similar To \\  Kinect V1} &                                    Color, Depth &                            Mirror Mask & \makecell[l]{7011 Scenes With \\ Mirror} \\
 \makecell[l]{Princeton \\ Tracking \\ Benchmark} &       \cite{song2013tracking} & 2013 &          Indoor &                                  Tracking (Other) &                     Structured Light &                                                                                                                         Kinect V1 &                                    Color, Depth &                                      - &                            100 Sequences \\
 \rowcolor{gray!25}                                   Dynamic Scene &        \cite{yoon2020dynamic} & 2020 & Indoor, Outdoor &     \makecell[l]{Novel View \\ Synthesis (Other)} &                                  SCS &                                                                                                                               Stereo Camera &                                           Color & \makecell[l]{Semantic \\ Segmentation} &                                 9 Scenes \\
                                       LightField & \cite{srinivasan2017learning} & 2017 & Other (Flowers) & \makecell[l]{Synthesizes A 4D \\ RGBD LF (Other)} &                                  SCS &                                                                                  \makecell[l]{Lytro Illum (Light \\ Field) (Stereo Camera)} &                                           Color &                                      - &                              3343 Images \\

 \bottomrule[2pt]
\end{NiceTabular}

}
\end{table*}

\subsection{Body}

In this category, all datasets are focused on body activities, such as action recognition, facial expression, hand activities, and sign language recognition. Here, we have only two sub-categories: the first one encompass full-body activities and the second one includes partial body parts, such as hands or face.

It is essential to notice that some of these datasets also include depth maps of the scene, but the focus of the dataset is on the Human Body (or part of it). Therefore, they are classified in this category.

\subsubsection{Human Activities}

This sub-category has all datasets focused on human activities, such as drinking, eating, playing tennis, and walking. Here, we have datasets that analyze actions for an individual person~\citep{wang2012mining, wang2014cross} or two-person interactions~\citep{6239234}.

The majority of the works in the ``Human Activities'' sub-category are collected in controlled scenes, and we only found  Hollywood 3D~\citep{hadfield2013hollywood} using in-the-wild datasets. The majority of the datasets are indoor scenes, but as they are centered on actions, they are classified in the ``Scene Type'' column as ``Full Body''. The most common extra data is the person pose (or skeleton) of the people involved in the scene. Such information can help improve automatic action recognition algorithms. Datasets containing Human Activities are presented in Table~\ref{TabelaHuman}. Next, we present three influential datasets in this sub-category: 
\newline

\begin{table*}[!t]

\caption{Datasets of ``Human Activity'' sub-category}
\label{TabelaHuman}
\resizebox{\textwidth}{!}{%

\begin{NiceTabular}{lllllllll}
\toprule[2pt]
                                                                           \textbf{Dataset Name} & \textbf{Ref.} & \textbf{Year} &     
\textbf{Scene Type} &  \textbf{Sensor Type} &                                              \textbf{Sensor Name} &                                                           \textbf{Data Modalities} &                                         \textbf{Extra Data} &                                                                                                          \textbf{Images/Scenes} \\
\midrule[1.5pt]

 \rowcolor{gray!25}
                                                  Depth 2 Height &           \cite{yin2020accurate} & 2020 &      Full Body &              TOF &                                                Kinect V2 &                                                              Color, Depth &                                                  - &                                                                                                            2136 Images \\

    SOR3D-AFF &            \cite{thermos2020deep} & 2020 &   Full Body &              TOF &                                                Kinect V2 &                                                              Color, Depth &                                                  - &                                                                1201 Sequences \\
    
        NTU RGB+D 120 &            \cite{liu2019ntu} & 2019 &   Full Body &              TOF &                                                Kinect V2 &                                                              Color, Depth, IR &                                                  \makecell[l]{Person Pose \\ (Skeleton)} &                                                                111480 Sequences \\
 \rowcolor{gray!25}
                        -- &            \cite{Tang_2019_ICCV} & 2019 &   Full Body &              TOF &                                                Kinect V2 &                                                              Color, Depth &                                                  - &                                                                \makecell[l]{800 Frames For Each \\ Person (26 People)} \\
    CMDFALL &            \cite{8546308} & 2018 &   Full Body &              Structured Light &                                                Kinect V1 &                                                              \makecell[l]{Color, Depth \\ Accelerometer} &                                                  - &                                                                20 Sequences \\
 \rowcolor{gray!25}

    UESTC &            \cite{ji2019large} & 2018 &   Full Body &              TOF &                                                Kinect V2 &                                                              Color, Depth &                                                  - &                                                                25600 Sequences \\
                          
                           \makecell[l]{UOW Online \\ Action3D} &            \cite{tang2018online} & 2018 &   Full Body &              TOF &                                                Kinect V2 &                                                              Color, Depth &            \makecell[l]{Person Pose \\ (Skeleton)} &                     \makecell[l]{20 Sequences (20 \\ Participants \\ Performing \\ Multiple Actions \\ In A Sequence)} \\
 \rowcolor{gray!25}             
     PKU-MMD &            \cite{liu2017pku} & 2017 &   Full Body &              TOF &                                                Kinect V2 &                                                              \makecell[l]{Color, Depth \\ Accelerometer} &                                                  \makecell[l]{Person Pose \\ (Skeleton)} &                                                                3076 Sequences \\

 TVPR &          \cite{10.1007/978-3-319-56687-0_1} & 2017 &   Full Body & Structured Light &                                      \makecell[l]{Asus Xtion\\ Pro Live} &                                                              Color, Depth &                                                  - &                                                                  \makecell[l]{23 Sequences (100 \\ People, 2004 Secs)} \\
  \rowcolor{gray!25}

    \makecell[l]{Chalearn LAP \\ IsoGD}  &            \cite{wan2016chalearn} & 2016 &   Full Body &              Structured Light &                                                Kinect V1 &                                                              Color, Depth &                                                  - &                                                                47933 Sequences  \\
 
    G3D  &            \cite{bloom2016hierarchical} & 2016 &   Full Body &              Structured Light &                                                Kinect V1 &                                                              Color, Depth & \makecell[l]{Person Pose \\ (Skeleton) \\ Semantic \\ Segmentation} &   \makecell[l]{7 Sequences \\ (Multiple Actions \\ Per Sequence)}  \\
 \rowcolor{gray!25}
  HHOI &                \cite{ShuIJCAI16} & 2016 &      Full Body &              TOF &                                                Kinect V2 &                                                              Color, Depth &            \makecell[l]{Person Pose \\ (Skeleton)} & \makecell[l]{8 Actors Recorded \\ Interections. \\ Each Interaction \\ Lasts 2-7 Seconds \\ Presented At 10-15 \\ Fps} \\

    \makecell[l]{ISR-UoL 3D \\ Social Activity}  &            \cite{Coppola2016a} & 2016 &   Full Body &              Structured Light &                                                Kinect V1 &                                                              Color, Depth &                                                  \makecell[l]{Person Pose \\ (Skeleton)} & 10 Sequences \\

 \rowcolor{gray!25}        NTU RGB+D &            \cite{shahroudy2016ntu} & 2016 &   Full Body &              TOF &                                                Kinect V2 &                                                              Color, Depth, IR &                                                  \makecell[l]{Person Pose \\ (Skeleton)} &                                                                56880 Sequences \\
                         \makecell[l]{TST Fall Detection \\ V2} &    \cite{gasparrini2015proposal} & 2016 &   Full Body &              TOF &                                                Kinect V2 &                              \makecell[l]{Color, Depth, \\ Accelerometer} &            \makecell[l]{Person Pose \\ (Skeleton)} &                                                                                                             264 Scenes \\
    \rowcolor{gray!25}        \makecell[l]{UOW LargeScale \\ Combined Action3D} &            \cite{zhang2016large} & 2016 &   Full Body &              TOF &                                                Kinect V2 &                                                              Color, Depth &            \makecell[l]{Person Pose \\ (Skeleton)} &                                                                                                         4953 Sequences \\

  CMU Panoptic &            \cite{Joo_2017_TPAMI} & 2015 &      Full Body &              TOF &                                                Kinect V2 &                                                              Color, Depth &                                        3D Skeleton &                                                                      \makecell[l]{65 Sequences (5.5 \\ Capture Hours)} \\
                     
   \rowcolor{gray!25}      SYSU 3D HOI &            \cite{hu2017jointly} & 2015 &   Full Body &              Structured Light &                                                Kinect V1 &                                                              Color, Depth &                                                  \makecell[l]{Person Pose \\ (Skeleton)} &                                                                480 Sequences \\
                     
                      \makecell[l]{TST Intake \\ Monitoring V1} & \cite{gasparrini2015performance} & 2015 &   Full Body & Structured Light &                                                Kinect V1 &                                                              Color, Depth &                                                  - &                                                                                                           48 Sequences \\
 \rowcolor{gray!25}                     \makecell[l]{TST Intake \\ Monitoring V2} & \cite{gasparrini2015performance} & 2015 &   Full Body & Structured Light &                                                Kinect V1 &                                                              Color, Depth &                                                  - &                                                                                                           60 Sequences \\
                                               TST TUG DataBase &        \cite{cippitelli2015time} & 2015 &   Full Body &              TOF &                                                Kinect V2 &                              \makecell[l]{Color, Depth, \\ Accelerometer} &            \makecell[l]{Person Pose \\ (Skeleton)} &                                                                                                           60 Sequences \\
 \rowcolor{gray!25}                                                      UTD-MHAD &               \cite{chen2015utd} & 2015 &   Full Body & Structured Light &                                                Kinect V1 &                              \makecell[l]{Color, Depth, \\ Accelerometer} &            \makecell[l]{Person Pose \\ (Skeleton)} &                                                                                                          861 Sequences \\
                                                      Human3.6M &                 \cite{h36m_pami} & 2014 &   Full Body &              TOF & \makecell[l]{MESA Imaging \\ SR4000 From \\ SwissRanger} &              \makecell[l]{Color, Depth, \\ Motion Capture \\ (mx) Camera} &            \makecell[l]{Person Pose \\ (Skeleton)} &                                                       \makecell[l]{447260 RGB-D \\ Frames (almost \\ 3.6M RGB Frames)} \\
 \rowcolor{gray!25}  KARD &            \cite{6990523} & 2014 &   Full Body &              Structured Light &                                                Kinect V1 &                                                              Color, Depth &                                   \makecell[l]{Person Pose \\ (Skeleton)} &                                                                540 Sequences \\

        LIRIS &            \cite{wolf2014evaluation} & 2014 &   Full Body &              Structured Light &                                                Kinect V1 &                                                              Color, Depth &                                                  - &                                                                180 Sequences \\
    \rowcolor{gray!25}  MAD &            \cite{huang2014sequential} & 2014 &   Full Body &              Structured Light &                                                Kinect V1 &                                                              Color, Depth &                                                  \makecell[l]{Person Pose \\ (Skeleton)} &                                                                40 Sequences \\
                                                      
  \makecell[l]{Northwestern-UCLA \\ Multiview Action 3D} &             \cite{wang2014cross} & 2014 &   Full Body & Structured Light &                                                Kinect V1 &                                                              Color, Depth &                                                  - &                                                                                                         1473 Sequences \\
 \rowcolor{gray!25} \makecell[l]{Online RGBD \\ Action Dataset \\ (ORGBD)} &            \cite{yu2014discriminative} & 2014 &   Full Body &              Structured Light &                                                Kinect V1 &                                                              Color, Depth &                                                  \makecell[l]{Person Pose \\ (Skeleton)} &                                                                48 Sequences \\
                        
                         \makecell[l]{TST Fall Detection \\ V1} &       \cite{gasparrini2014depth} & 2014 &   Full Body & Structured Light &                                                Kinect V1 &                              \makecell[l]{Color, Depth, \\ Accelerometer} &            \makecell[l]{Person Pose \\ (Skeleton)} &                                                                                                           20 Sequences \\
 \rowcolor{gray!25} UR Fall Detection &           \cite{kwolek2014human} & 2014 &      Full Body & Structured Light &                                                Kinect V1 &                              \makecell[l]{Color, Depth, \\ Accelerometer} &                                                  - &                                                                                                           70 Sequences \\                         
                         
\makecell[l]{Chalearn \\ Multimodal \\ Gesture \\ Recognition} &         \cite{escalera2013multi} & 2013 &   Full Body & Structured Light &                                                Kinect V1 &                                      \makecell[l]{Color, Depth, \\ Audio} & \makecell[l]{User Mask, Person \\ Pose (Skeleton)} &                                                                        \makecell[l]{707 Sequences \\ (1720800 Frames)} \\

   \rowcolor{gray!25}   \makecell[l]{Florence 3D \\ Actions} &            \cite{6595917} & 2013 &   Full Body &              Structured Light &                                                Kinect V1 &                                                              Color, Depth &                                   \makecell[l]{Person Pose \\ (Skeleton)} &                                                                215 Sequences \\

 \bottomrule[2pt]

\end{NiceTabular}

}

 \vspace{-0.3cm}
 \begin{flushright}\scriptsize Continue on Next Page \end{flushright}        

\end{table*}

\begin{table*}[!t]

\resizebox{\textwidth}{!}{%

\begin{NiceTabular}{lllllllll}
\toprule[2pt]
                                                                           \textbf{Dataset Name} & \textbf{Ref.} & \textbf{Year} &     
\textbf{Scene Type} &    \textbf{Sensor Type} &                                            \textbf{Sensor Name} &                                                           \textbf{Data Modalities} &                                         \textbf{Extra Data} &                                                                                                          \textbf{Images/Scenes} \\
\midrule[1.5pt]

 \rowcolor{gray!25}

        IAS-Lab RGBD-ID &            \cite{munaro20133d} & 2013 &   Full Body &              Structured Light &                                                Kinect V1 &                                                              Color, Depth &                                   \makecell[l]{Person Pose \\ (Skeleton) \\ Semantic \\ Segmentation} &                                                                33 Sequences \\
        
MHAD &          \cite{ofli2013berkeley} & 2013 &   Full Body & Structured Light &                                                Kinect V1 & \makecell[l]{Color, Depth, \\ Accelerometer, \\ Motion Capture \\ System} &                                                  - &                                                                                                          660 Sequences \\

\rowcolor{gray!25}  Mivia Action &            \cite{Carletti13} & 2013 &   Full Body &              Structured Light &                                                Kinect V1 &                                                              Color, Depth &                                   - &                                                                28 Sequences \\

\makecell[l]{ChaLearn Gesture \\ Challenge} &         \cite{guyon2014chalearn} & 2012 &   Full Body & Structured Light &                                                Kinect V1 &                                                              Color, Depth &                                                  - &                                                                                                        50k Sequences \\
 
 \rowcolor{gray!25}              DGait &           \cite{borras2012depth} & 2012 &   Full Body & Structured Light &                                                Kinect V1 &                                                              Color, Depth &                                                  - &                                                                           \makecell[l]{583 Sequences (53 \\ Subjects)} \\

 \makecell[l]{MSR \\ DailyActivity3D} &            \cite{wang2012mining} & 2012 &   Full Body & Structured Light &                                                Kinect V1 &                                                              Color, Depth &            \makecell[l]{Person Pose \\ (Skeleton)} &                                                                                                          320 Sequences \\
 \rowcolor{gray!25} RGBD-ID &            \cite{Barbosa:reid12} & 2012 &   Full Body & Structured Light &                                                Kinect V1 &                                                              Color, Depth &            \makecell[l]{Person Pose \\ (Skeleton)} &                                                                             \makecell[l]{316 Sequences (79 \\ People)} \\
\makecell[l]{SBU Kinect \\ Interaction} &             \cite{6239234} & 2012 &   Full Body & Structured Light &                                                Kinect V1 &                                                              Color, Depth &            \makecell[l]{Person Pose \\ (Skeleton)} &                                                               \makecell[l]{21 Sequences From \\ Seven \\ Participants} \\
 \rowcolor{gray!25} UTKinect-Action3D &            \cite{xia2012view} & 2012 &   Full Body &              Structured Light &                                                Kinect V1 &                                                              Color, Depth &                                   \makecell[l]{Person Pose \\ (Skeleton)} &                                                                200 Sequences \\
RGB-D People &                   \cite{6095074} & 2011 &      Full Body & Structured Light &                                                Kinect V1 &                                                              Color, Depth &     \makecell[l]{Object Detection \\ And Tracking} &                                                              \makecell[l]{1 Sequence (1132 \\ Frames Of 3 \\ Sensors)} \\                       
 \rowcolor{gray!25} MSR Action3D &                   \cite{5543273} & 2010 &   Full Body & Structured Light &               \makecell[l]{Similar to \\ Kinect V1 (N/A)} &                                                              Color, Depth &                                                  - &                                                                          \makecell[l]{557 Sequences \\ (23797 Frames)} \\
 Hollywood 3D &     \cite{hadfield2013hollywood} & 2013 & In-the-wild &              SCS &                                                      Stereo Camera &                                                              Color, Depth &                                                  - &                                                                                \makecell[l]{Around 650 Video \\ Clips} \\

 \bottomrule[2pt]

\end{NiceTabular}

}
\end{table*}

\textbf{NTU RGB+D.} This dataset contains more than 50,000 video samples representing 60 distinct actions that are divided into three major groups: health-related actions (e.g., falling down, staggering), 40 daily actions (e.g., eating, drinking), 11 mutual actions (e.g., kicking, hugging)~\citep{shahroudy2016ntu}. Forty subjects aged between 10 and 35 performed the actions in this dataset. The dataset was collected using three Kinect~v2 from different horizontal views and is available with RGB, Depth, infrared (IR) sequences, and person pose (skeleton) information.

The authors extended the NTU RGB+D to a new dataset called NTU RGB+D 120, which contains other 60 classes and 57,600 samples, also containing the same capturing system and data modalities as the previous dataset~\citep{liu2019ntu}.

\textbf{MSR DailyActivity3D Dataset.} This dataset covers sixteen different activities: drink, eat, read a book, call cellphone, write on a paper, use a laptop, use a vacuum cleaner, cheer up, sit still, toss paper, play games, lie down on a sofa, walk, play guitar, stand up, and sit down~\citep{wang2012mining}. Ten subjects performed each action twice: one for standing and one for sitting position. This dataset also includes person pose information for each frame. The authors used the Kinect~v1 to acquire the depth of the scenes.
\newline

\textbf{MSR Action3D.} This dataset covers twenty different actions performed by ten subjects. Each action was performed two to three times, resulting in 557 filtered sequences and 23,797 frames~\citep{5543273}. The actions are divided into three sets, where the first categorize actions with similar moviments. The third set is composed by complex actions together. All sequences were acquired using Kinect~v1 sensor.

\subsubsection{Gestures (Partial Body)}

Here, we grouped all works that involve human actions or activities and have data available for human body parts, such as arms, head, and hand. There is a wide variety of dataset purposes in this sub-category, such as action recognition based on a first-person view (no torso/head parts available in video)~\citep{tang2017action}, salad preparation~\citep{stein2013combining}, hand-pose information~\citep{tompson14tog}, and sign language recognition~\citep{wang2012robust}.

\begin{table*}[!t]

\caption{Datasets of ``Gestures (Partial Body) sub-category''}
\label{TabelaGestures}
\resizebox{\textwidth}{!}{%

\begin{NiceTabular}{lllllllll}
\toprule[2pt]
                                                                           \textbf{Dataset Name} & \textbf{Ref.} & \textbf{Year} &     
\textbf{Scene Type} &  \textbf{Sensor Type} &                                              \textbf{Sensor Name} &                                                           \textbf{Data Modalities} &                                         \textbf{Extra Data} &                                                                                                          \textbf{Images/Scenes} \\
\midrule[1.5pt]

 \rowcolor{gray!25}
   \makecell[l]{Bimanual \\ Actions} &                  \cite{dreher2020learning} & 2020 & \makecell[l]{Part Of Body (hand, \\ Head, Etc.)} &                            Structured Light &            \makecell[l]{PrimeSense \\ Carmine 1.09} &                                                     Color, Depth & - &            \makecell[l]{540 Sequences} \\

                                    RGB2Hands &                  \cite{wang_SIGAsia2020} & 2020 & \makecell[l]{Part Of Body (hand, \\ Head, Etc.)} &                            Structured Light &            \makecell[l]{Intel RealSense \\ SR300/Synthetic} &                                                     Color, Depth & \makecell[l]{Segmentation, 2D \\ Keypoints, Dense \\ Matching Map, \\ Inter-hand \\ Distance, \\ Intra-hand \\ Distance} &            \makecell[l]{Real: 4 Sequences \\ (1724 Frames). \\ Synthetic: NA} \\
 \rowcolor{gray!25}                                       ObMan &                    \cite{hasson19_obman} & 2019 & \makecell[l]{Part Of Body (hand, \\ Head, Etc.)} &                                           - &                                                   Synthetic &                                                     Color, Depth &                                         \makecell[l]{3D Hand Keypoints, \\ Object \\ Segmentation, \\ Hand Segmentation} &                                                                 150k Images \\
  EgoGesture &                  \cite{zhang2018egogesture} & 2018 & \makecell[l]{Part Of Body (hand, \\ Head, Etc.)} &                            Structured Light &            \makecell[l]{Intel RealSense \\ SR300/Synthetic} &                                                     Color, Depth & - &            \makecell[l]{2081 Sequences} \\
 \rowcolor{gray!25}                                               -- &        \cite{8578148} & 2018 & \makecell[l]{Part Of Body (hand, \\ Head, Etc.)} &                            Structured Light &                      \makecell[l]{Intel RealSense \\ SR300} & \makecell[l]{Color, Depth, \\ Magnetic And \\ Kinematic Sensors} &                                                                                                                        - &                       \makecell[l]{1175 Sequences \\ (over 100k \\ Frames)} \\
                              BigHand2.2M &                    \cite{Yuan_CVPR_2017} & 2017 & \makecell[l]{Part Of Body (hand, \\ Head, Etc.)} &                            Structured Light &                      \makecell[l]{Intel RealSense \\ SR300} &                \makecell[l]{Color, Depth, 6D \\ Magnetic Sensor} &                                                                                                                        - &            \makecell[l]{N/A Sequences (2.2 \\ Million Images), \\ 10 Subjects} \\
  \rowcolor{gray!25}                                         Pandora &                \cite{borghi2017poseidon} & 2017 & \makecell[l]{Part Of Body (hand, \\ Head, Etc.)} &                                         TOF &                                                   Kinect V2 &                                                     Color, Depth &                                                               \makecell[l]{Upper Body Part \\ Person Pose \\ (Skeleton)} & \makecell[l]{100 Sequences \\ (more Than 250k \\ Frames) From 20 \\ Subjects} \\
                                       RHD &                        \cite{zimmermann2017learning} & 2017 & \makecell[l]{Part Of Body (hand, \\ Head, Etc.)} &                                           - &                                                   Synthetic &                                                     Color, Depth &                                                                                 \makecell[l]{Segmentation, \\ Keypoints} &                                                                  43986 Images \\
   \rowcolor{gray!25}                                       THU-READ &                    \cite{tang2017action} & 2017 & \makecell[l]{Part Of Body (hand, \\ Head, Etc.)} &                            Structured Light &                         \makecell[l]{PrimeSense \\ Carmine} &                                                     Color, Depth &                                                                                                                        - &                                                                1920 Sequences \\
                                      STB &                     \cite{zhang2017hand} & 2016 & \makecell[l]{Part Of Body (hand, \\ Head, Etc.)} &      \makecell[l]{SCS, Structured \\ Light} &                 \makecell[l]{Intel Real Sense \\ F200, Stereo Camera} &                                                     Color, Depth &                                                                                                                        - &                                  \makecell[l]{12 Sequences \\ (18k Images)} \\
  \rowcolor{gray!25}                              
  Creative Senz3D &                  \cite{memo2015exploiting, memo2018head} & 2015 & \makecell[l]{Part Of Body (hand, \\ Head, Etc.)} &                            Structured Light &            Creative Senz3D &                                                     Color, Depth & - &            \makecell[l]{1320 Sequences} \\
                                      EYEDIAP &                  \cite{funes2014eyediap} & 2014 & \makecell[l]{Part Of Body (hand, \\ Head, Etc.)} &                            Structured Light &                                                   Kinect V1 &                                                     Color, Depth &                                                                                   \makecell[l]{Eye Points, Head \\ Pose} &                                                                  94 Sequences \\
 \rowcolor{gray!25}        \makecell[l]{Eurecom Kinect \\ Face} &                  \cite{IEEETransactions} & 2014 & \makecell[l]{Part Of Body (hand, \\ Head, Etc.)} &                            Structured Light &                                                   Kinect V1 &                                                     Color, Depth &                                                                                                              Face Points &                                                                 936 Sequences \\
  Hand Gesture &                  \cite{marin2014hand, marin2016hand} & 2014 & \makecell[l]{Part Of Body (hand, \\ Head, Etc.)} &                            Structured Light &            Kinect V1 &                                                     Color, Depth & - &            \makecell[l]{1400 Sequences} \\
          \rowcolor{gray!25}                                  MANIAC &                    \cite{aksoy2015model} & 2014 & \makecell[l]{Part Of Body (hand, \\ Head, Etc.)} &                            Structured Light &                                                   Kinect V1 &                                                     Color, Depth &                                                                                                                        - &                                                                 103 Sequences \\
                               NYU Hand Pose &                      \cite{tompson14tog} & 2014 & \makecell[l]{Part Of Body (hand, \\ Head, Etc.)} &                            Structured Light &                        \makecell[l]{Synthetic/Kinect \\ V1} &                                                     Color, Depth &                                                                                                                Hand Pose &                                                                  81009 Frames \\
 \rowcolor{gray!25}                                            3DMAD &                 \cite{nesli2013spoofing} & 2013 & \makecell[l]{Part Of Body (hand, \\ Head, Etc.)} &                            Structured Light &                                                   Kinect V1 &                                                     Color, Depth &                                                                                                               Eye Points &                                 \makecell[l]{255 Sequences \\ (76500 Frames)} \\

                                   50 Salads &                \cite{stein2013combining} & 2013 & \makecell[l]{Part Of Body (hand, \\ Head, Etc.)} &                            Structured Light &                                                   Kinect V1 &                     \makecell[l]{Color, Depth, \\ Accelerometer} &                                                                                 \makecell[l]{Activity \\ Classification} &                                     \makecell[l]{50 Sequences (25 \\ People)} \\

 \rowcolor{gray!25}                                    Dexter 1 &              \cite{handtracker_iccv2013} & 2013 & \makecell[l]{Part Of Body (hand, \\ Head, Etc.)} & \makecell[l]{SCS, Structured \\ Light, TOF} & \makecell[l]{Kinect V1, Creative\\  Gesture Camera,\\ Stereo Camera} &                                                     Color, Depth &                                                                                                                        - &                                                                   7 Sequences \\
                                           -- &                            \cite{xu2013} & 2013 & \makecell[l]{Part Of Body (hand, \\ Head, Etc.)} &                                         TOF &                                          SoftKinetic DS 325 &            \makecell[l]{Color, Depth, \\ Measurand \\ ShapeHand} &                                                                                                                        - &                                     \makecell[l]{870 Images (30 \\ Subjects)} \\
 \rowcolor{gray!25}\makecell[l]{MSRC-12 \\ Kinect Gesture} &                           \cite{10.1145/2207676.2208303} & 2012 & \makecell[l]{Part Of Body (hand, \\ Head, Etc.)} &                            Structured Light &                                                   Kinect V1 &                                                     Color, Depth &                                                                                                                        - &                                \makecell[l]{594 Sequences \\ (719359 Frames)} \\
                                MSR Gesture3D &                    \cite{wang2012robust} & 2012 & \makecell[l]{Part Of Body (hand, \\ Head, Etc.)} &                            Structured Light &                                                   Kinect V1 &                                                     Depth &                                                                                                                        - &                                                                 336 Sequences \\
 \rowcolor{gray!25}                           Florence 3D Faces & \cite{Bagdanov:2011:FHF:2072572.2072597} & 2011 & \makecell[l]{Part Of Body (hand, \\ Head, Etc.)} &                                           - &                                                   Synthetic &                                                            Color &                                                                                                                        - &                                   \makecell[l]{53 People (N/A \\ Frames/Seqs)} \\

 \bottomrule[2pt]
\end{NiceTabular}

}
\end{table*}
The most cited datasets in this sub-category include:
\newline

\textbf{NYU Hand Pose Dataset.} This dataset was captured using three Kinect~v1, with two side views and a frontal view. The authors also re-created a synthetic hand pose for each view~\citep{tompson14tog}, and made available 36 hand point locations for each frame. Three people acquired the data: one person used for training and the other two for testing, leading to over 80 thousand acquired frames.
\newline

\textbf{MSR Gesture3D.} This dataset contains sign language gestures. The authors collected 12 dynamic American Sign Language (ASL) gestures from ten people. The dataset was captured using Kinect~v1, and has 336 sequences since each person performed multiple recordings of all selected signs. The authors performed a hand segmentation, and depth information is available only for the segmented hand regions. Background and body portions below the wrist were removed.

\subsection{Medical}

In this category, we present datasets that are from any part of the medical field. The exclusion criteria removed most of the datasets found here because these contained only private data. For instance, we collected eleven datasets containing endoscopic data, but only three meets all criteria to be included in our work. This situation is common in medical applications as sharing medical information requires regulated procedures.

We found only four datasets available in this category, of which three of them contain endoscopic data and one contains 3D models of the iris. The most cited dataset in containing depth information in the medical field is:
\newline

\textbf{Colonoscopy CG Dataset.} This dataset is composed of endoscopic data of the colon. To the best of our knowledge, this is the most frequent type of data that contains depth maps in the Medical category, even if analyzing datasets with non-shared data. The authors generated a synthetic dataset using Unity graphic engine based on a human CT colonography scan. They extracted a surface mesh using manual segmentation and meshing~\citep{rau2019implicit}. Their work also proposed and tested an algorithm in real data, but this data is not available for the community thus not included in this paper.

\begin{table*}[!t]

\caption{Datasets of ``Medical'' Category.}
\label{TabelaMedical}
\resizebox{\textwidth}{!}{%

\begin{NiceTabular}{lllllllll}
\toprule[2pt]
                                                                           \textbf{Dataset Name} & \textbf{Ref.} & \textbf{Year} &     
\textbf{Scene Type} &  \textbf{Sensor Type} &                                              \textbf{Sensor Name} &                                                           \textbf{Data Modalities} &                                         \textbf{Extra Data} &                                                                                                          \textbf{Images/Scenes} \\
\midrule[1.5pt]

 \rowcolor{gray!25}
          SCARED &  \cite{allan2021stereo} & 2021 &  Endoscopy & \makecell[l]{SCS, Structured \\ Light} & \makecell[l]{Structured Light \\ System (using P300 \\ Neo Pico), Stereo Camera} &    Color, Depth &          - &                                9 Sequences \\
   Colonoscopy CG &  \cite{rau2019implicit} & 2019 &  Endoscopy &                                      - &                                                              Synthetic &    Color, Depth &          - &                               16016 Images \\
 \rowcolor{gray!25}Endoscopic Video &          \cite{mountney2010three} & 2010 &  Endoscopy &                                    SCS &                                                                    Stereo Camera &           Color &          - &                                  25 Scenes \\
               -- & \cite{benalcazar20203d} & 2020 &    Iris Scan & - &                                                             Synthetic &    Color, Depth &          - & \makecell[l]{100 Irises (72k \\ Images)} \\

 \bottomrule[2pt]
\end{NiceTabular}

}
\end{table*}

\section{Discussion}
\label{sec:discussion}

The datasets presented in Section~\ref{sec:datasets} compose a collection of different scenes, sensors, and activities. We provide information about the Sensor Type, Number of Images/Scenes, Scene Type, Sensor Name, and Data Modalities available for each dataset. Unlike previous surveys of RGB-D datasets~\citep{firman2016rgbd}, we do not categorize the datasets regarding their realism since this is a subjective criterion and it is up to the researcher who will analyze the datasets to decide. Despite the variety of datasets presented, we identified common tendencies in all areas and discussed them in this section.

Although synthetic data is becoming more present each time, the usage of real data is presented in the majority of the datasets. Comparing the 2016-2018 to the 2019-2021 trienniums, we found a 50\% increase in the numbers of datasets containing synthetic data. Synthetic datasets are usually cheaper to produce than performing real data acquisition because extra annotations, e.g., semantic segmentation or object tracking, are automatically generated. On the other hand, complex scene annotations for real data are costly, especially in scenes such as driving and aerial.

Synthetic datasets were initially created using simulators~\citep{tarel2010improved, tarel2012vision}, but these simulators were distinct to real-world scenarios since the computational power of the machines was limited. Hence, it was not possible to generate consistent and realistic datasets for complex scenes. Recently, realistic simulators were created for driving scenes, such as CARLA~\citep{Dosovitskiy17}, Nvidia Drive Sim\footnote{\url{https://developer.nvidia.com/drive/drive-sim}}, and indoor scenes, such as Habitat~\citep{szot2021habitat, habitat19iccv}. Despite the usage of simulators, other datasets rely on game engines or general computer graphics engines to build their systems, such as SYNTHIA~\citep{Ros_2016_CVPR}, Virtual KITTI~\citep{Gaidon:Virtual:CVPR2016}, and Virtual KITTI 2~\citep{cabon2020vkitti2} that used Unity\footnote{\url{https://unity.com/}} as graphic engine, and GTA-SfM~\citep{wang2020flow} that uses scenes from the game GTAV.

The usage of synthetic data has been combined with real data to produce more complex scenes. These are applied especially for techniques that explore the generalization of their methods in non-expected scenes, i.e., using datasets not used in the training step~\citep{ummenhofer2017demon, Ranftl2020, eftekhar2021omnidata}. 

These papers combine datasets containing different types of acquisition and scenes to produce generalizable models. \citet{Ranftl2020} created multiple cross-dataset training strategies, and its combination of datasets with more images ---called MIX5--- contains data from DIML, MegaDepth, RedWeb, WSVD, and 3D Movies datasets. \citet{ranftl2021vision} expanded this combination, creating the MIX6 cross-dataset set containing about 1.4 million training images. Both works were evaluated using a mixture of testing datasets. The robustness of the models are also evaluated in a cross-dataset strategy for estimating depth from a monocular video~\citep{kopf2021robust}, and instead of testing in multiple types of scenes, \citet{ji2021monoindoor} combined distinct datasets of the same type of scene to improve the results for the indoor environment.

Recently, domain adaptation has been applied to improve the performance of the combination of datasets in the training step~\citep{guo2018learning, atapour2018real, zhao2019geometry}.~\citet{atapour2018real}, for instance, combines one synthetic and one real dataset using domain adaptation to improve the result of training. They claim that directly using synthetic data may not improve the results for realistic data evaluation due to dataset bias. They adapt the domain of a synthetic dataset to a real dataset using Style Transfer and combine them to train their models.~\citet{zhao2019geometry} also performs domain adaptation, and they claim that due to the lack of paired synthetic and real images, the synthetic-to-realistic image translation adds distortions to the depth estimation. They overcome this difficulty by exploring a more complex training procedure involving synthetic-to-realistic and realistic-to-synthetic translations. To generate more realistic synthetic data,~\citet{7410665} proposed the use of 3D CAD Models to produce 2D synthetic images, since these CAD Models allow multiple viewpoints and complete control of the deformations in the modeled objects to increase the variability of the created dataset.~\citet{8374552} also used 3D CAD Models, but they intended to create realistic depth data from the 3D objects. They proposed a framework that simulates real distortion factors of depth data acquisition, e.g., material reflectance and sensor noise, to generate reliable depth data.  
In addition to using synthetic data, domain adaptation could also be applied to real-to-real translation~\citep{lopez2020desc, hornauer2021visual} since the dataset bias also affects distinct real datasets, especially by variations of scale and capture's position of the scenes~\citep{torralba2011unbiased}.


\section{Conclusions}
\label{sec:conclusion}

In this work, we presented a survey of publicly available image datasets that contain depth information. We categorized and summarized over 200 datasets based on the image scenes, sensors used to collect the depth information, and the different applications for which these datasets can be used. Almost half of the datasets we describe were proposed after the publication of the last survey~\citep{cai2017rgb}. The new datasets expand the scope of applications that depth datasets can be used for, such as medical applications. The new datasets also expand the quality and quantity of data for other areas.

We also presented different forms of acquiring depth information from a scene. We expect that this explanation could be used in conjunction with extra information of the datasets to allow researchers to choose the ones that best fulfill their needs. Researchers of zero-shot learning trying to increase generalization capabilities for their model could also benefit from our work since they may select distinct datasets in terms of sensor type, application, and scene type for training and evaluating their methods.

\section*{CRediT authorship contribution statement}

{\bf Alexandre Lopes:} Conceptualization, Formal analysis, Investigation, Methodology, Writing - review \& editing. {\bf Roberto Souza:} Funding acquisition, Methodology, Project administration, Supervision, Writing -- review \& editing. {\bf Helio Pedrini:} Methodology, Project administration, Supervision, Writing -- review \& editing.

\section*{Declaration of competing interest}

The authors declare that they have no known competing financial interests or personal relationships that could have appeared to influence the work reported in this paper.

\section*{Acknowledgements}

The authors are grateful to the National Council for Scientific and Technological Development, Brazil (CNPq grant 309330/2018-1). Roberto Souza thanks the Natural Sciences and Engineering Research Council (NSERC - RGPIN-2021-02867) for ongoing operational support.

\bibliographystyle{model2-names}
\bibliography{refs}

\end{document}